\documentclass[journal,final]{sty/ieee/IEEEtran}
\usepackage[T1]{fontenc}
\usepackage[utf8x]{inputenc}

\newif\ifpeerreview
\newif\ifpeerreviewlines
\peerreviewlinestrue  
\makeatletter
\@ifclasswith{sty/ieee/IEEEtran}{peerreview}{\peerreviewtrue}{\peerreviewfalse\peerreviewlinesfalse}
\makeatother

\newcommand{\nodiffshere}{%
\ifpeerreviewlines
  \usepackage[switch]{lineno}
  \newcommand{\multicolslines}{%
    \linenumbers
    \def\makeLineNumber{\docolaction{\makeLineNumberLeft}{}{\makeLineNumberRight}}}
\else
  \newcommand{\multicolslines}{}
\fi
}
\nodiffshere

\newif\ifnature
\naturetrue

\usepackage[linesnumbered,commentsnumbered,ruled,vlined]{algorithm2e}  
\usepackage{amsmath}
\usepackage{amssymb}
\usepackage{balance}
\usepackage{booktabs}
\usepackage{cite}
\usepackage{color}
\usepackage{courier}
\usepackage{./sty/easygraphic}
\usepackage[obeyFinal]{easy-todo}
\usepackage{etextools}
\usepackage{etoolbox}
\usepackage{float}
\usepackage{hyperref}
\usepackage{ifdraft}
\usepackage{listings}
\usepackage{mathtools}
\usepackage{mdframed}
\usepackage[colaction]{multicol}
\usepackage{pdfpages}
\usepackage{pgf}
\usepackage{pgfkeys}
\usepackage{pifont}
\usepackage[binary-units=true, group-digits=integer, group-separator={,}, group-minimum-digits=4]{siunitx}
\usepackage{subcaption}
\usepackage{tabularx}  
\usepackage{xcolor}
\usepackage{xparse}
\usepackage{xstring}

\usepackage[capitalise]{cleveref}

\Crefname{paragraph}{Paragraph}{Paragraphs}

\makeatletter

\renewcommand{\todo}[2]{%
  \protect\@ifnextchar\bgroup{\protect\todoInner{#1}{#2}}{\protect\todoInner{#1}{#2}{}}
}
\newcommand{\todoInner}[3]{%
  \protect\todoii{~#1 #2 #3}{#1 #2}%
}
\newcommand\note[2]{\ifoptionfinal{}{\textcolor{orange}{#1 #2}}}
\makeatother

\def\figureFontSizeName{footnotesize}
\def\figureFontSize{\csname \figureFontSizeName \endcsname}
\SetAlCapFnt{\rm\figureFontSize}

\def\etal{\mbox{\em et al.}}

\DeclareDocumentCommand{\hanging}{O{1em} O{0em} +m}{%
    \begin{list}
            {}
            {\setlength{\itemindent}{-#1}%
                \setlength{\leftmargin}{#2+#1}%
                \setlength{\itemsep}{0pt}%
                \setlength{\parsep}{\parskip}%
                \setlength{\topsep}{\parskip}%
            }
    \setlength{\parindent}{-#1}%
    \item[]
    #3
    \end{list}
}

\pgfkeys{/fig/.is family, /fig,
    pos/.estore in=\figPos,
    label/.estore in=\figLabel,
    lof/.estore in=\figLof,
    default/.style = {
        pos=t,
        label=,
        lof=,
    }
}
\DeclareDocumentCommand{\fig}{s O{} +m +m}{%
    \pgfkeys{/fig, default, #2}

    \edef\figPoss{[\figPos]}
    \IfBooleanTF{#1}{
          \def\figDecl{\begin{figure*}}
          \def\figDeclEnd{\end{figure*}}
        }{
          \def\figDecl{\begin{figure}}
          \def\figDeclEnd{\end{figure}}
        }
    \expandaftercmds{\figDecl}{\figPoss}
        \centering
        {%
            \figureFontSize
            #3%
        }
        \ifdefempty{\figLof}{%
            \caption{#4}}{%
            \caption[\figLof]{#4}}
        \ifdefempty{\figLabel}{}{\label{\figLabel}}
    \figDeclEnd
}

\pgfkeys{/subfig/.is family, /subfig,
    caption/.estore in=\subfigCaption,
    label/.estore in=\subfigLabel,
    width/.estore in=\subfigWidth,
    default/.style = {
        caption={},  
        width=,
        label=,
    }
}
\DeclareDocumentCommand{\subfig}{O{} +m}{%
    \pgfkeys{/subfig, default, #1}
    \begin{subfigure}[t]{\ifdefempty{\subfigWidth}{\linewidth}{\subfigWidth}}
        \centering
        #2%
        \caption{\subfigCaption}
        \ifdefempty{\subfigLabel}{}{\label{\subfigLabel}}
    \end{subfigure}
}

\pgfkeys{/alg/.is family, /alg,
    label/.estore in=\algLabel,
    pos/.estore in=\algPos,
    default/.style = {
        label=,
        pos=t,
    }
}
\makeatletter
\newcommand{\algRemovelatexerror}{\let\@oldlatex@error\@latex@error \let\@latex@error\@gobble}
\newcommand{\algRestorelatexerror}{\let\@latex@error\@oldlatex@error}
\makeatother
\newlength\algboldlen
\DeclareDocumentCommand{\alg}{O{} +m +m}{%
    \pgfkeys{/alg, default, #1}

    \edef\algPoss{[\algPos]}
    \expandaftercmds{\begin{figure}}{\algPoss}
        \centering
        \algRemovelatexerror \begin{algorithm}[H] \algRestorelatexerror
            \caption{\figureFontSize #2}
            \ifdefempty{\algLabel}{}{\label{\algLabel}}

            \newcommand\BoldKw[2]{
                \settowidth\algboldlen{{\bf ##1: }}
                \begin{minipage}{\linewidth-\leftmargin+\itemindent}%
                    {\bf ##1:} \hanging[1.5em][\algboldlen]{\vspace{-1em} ##2}
                \end{minipage}}
            \newcommand\Input[1]{\BoldKw{Input}{##1}}
            \newcommand\Output[1]{\BoldKw{Output}{##1}}

            \figureFontSize #3
        \end{algorithm}
    \end{figure}
}

\pgfkeys{/wtable/.is family, /wtable,
    inline/.estore in=\wtableInline,
    padh/.estore in=\wtablePadh,
    pos/.estore in=\wtablePos,
    label/.estore in=\wtableLabel,
    lof/.estore in=\wtableLof,
    default/.style = {
        pos=t,
        label=,
        lof=,
        padh=,
        inline=,
    }
}
\DeclareDocumentCommand{\wtable}{O{} +m +m +m +m}{%
    \pgfkeys{/wtable, default, #1}

    \ifcsname savedPadh\endcsname%
    \else
      \newlength{\savedPadh}
    \fi
    \setlength{\savedPadh}{\tabcolsep}
    \ifdefempty{\wtablePadh}{}{%
        \setlength{\tabcolsep}{\wtablePadh}
    }

    \edef\wtablePoss{[\wtablePos]}
    \ifdefempty{\wtableInline}{
          \def\tabDecl{%
            \expandaftercmds{\begin{table}}{\wtablePoss}}
          \def\tabDeclEnd{\end{table}}
        }{
          \def\tabDecl{%
            \onecolumn
            \figureFontSize}
          \def\tabDeclEnd{\twocolumn}
        }
    {%
      \tabDecl
          \centering
          \ifdefempty{\wtableLof}{%
              \caption{#2}}{%
              \caption[\wtableLof]{#2}}
          \begin{tabularx}{\linewidth}{#3}
          \toprule
          #4
          \bottomrule
          \end{tabularx}
          \caption*{#5}

          \ifdefempty{\wtableLabel}{}{\label{\wtableLabel}}
      \tabDeclEnd
    }

    \setlength{\tabcolsep}{\savedPadh}
}
\newcommand{\valImagenetCohenAuc}{\SI{0.0022}{}}
\newcommand{\valCifarCohenAuc}{\SI{0.0065}{}}
\newcommand{\valCifarMadryAuc}{\SI{0.0124}{}}  
\newcommand{\valCifarMadryReproAuc}{\SI{0.0104}{}}  

\newcommand{\valCifarOurAucTraditional}{\SI{0.0013}{}}
\newcommand{\valCifarOurBtrAucTraditional}{\SI{0.0014}{}}

\newcommand{\valImagenetOurAucTraditional}{\SI{0.0004}{}}
\newcommand{\valImagenetOurBtrAucTraditional}{\SI{0.0013}{}}

\newcommand{\valImagenetOurAucModified}{\SI{0.0053}{}}

\newcommand{\valImagenetOurBenefitAuc}{\SI{2.4}{\times}}

\makeatletter
\newcommand{\wsecheading}[1]{\textbf{\boldmath #1}}
\newcommand{\wsubsecheading}[1]{\textbf{\boldmath #1}}
\let\Oldsection\section
\newcommand{\Newsection}[1]{\vspace{-0pt}\Oldsection{\wsecheading{#1}}\vspace{-0pt}}
\renewcommand{\section}{\@ifstar{\Oldsection*}\Newsection}
\let\Oldsubsection\subsection
\renewcommand{\subsection}[1]{\vspace{-0pt}\Oldsubsection{\wsubsecheading{#1}}\vspace{-0pt}}
\let\Oldsubsubsection\subsubsection
\renewcommand{\subsubsection}[1]{\vspace{-0pt}\Oldsubsubsection{\hspace*{4pt}\wsubsecheading{#1}}\vspace{-0pt}}

\makeatother

\author{%
    Walt Woods, Jack Chen, and Christof Teuscher \\
    Department of Electrical and Computer Engineering \\
    Portland State University, Portland, OR, USA \\
    \{wwoods, chenjac, teuscher\}@pdx.edu}

\title{Adversarial Explanations for Understanding Image Classification Decisions and Improved Neural Network Robustness \vspace{-0.0em}}

\begin{document}
\maketitle

\ifpeerreviewlines
  \linenumbers
\fi

\begin{abstract}%
\noindent
For sensitive problems, such as medical imaging or fraud detection, {\em Neural Network} (NN) adoption has been slow due to concerns about their reliability, leading to a number of algorithms for explaining their decisions.  NNs have also been found vulnerable to a class of imperceptible attacks, called adversarial examples, which arbitrarily alter the output of the network.
Here we demonstrate both that these attacks can invalidate prior attempts to explain the decisions of NNs, and that with very robust networks, the attacks themselves may be leveraged as explanations with greater fidelity to the model.
We show that the introduction of a novel regularization technique inspired by the Lipschitz constraint, alongside other proposed improvements, greatly improves an NN's resistance to adversarial examples.  On the ImageNet classification task, we demonstrate a network with an {\em Accuracy-Robustness Area} (ARA) of \valImagenetOurAucModified, an ARA \valImagenetOurBenefitAuc\ greater than the previous state of the art.
Improving the mechanisms by which NN decisions are understood is an important direction for both establishing trust in sensitive domains and learning more about the stimuli to which NNs respond.

\end{abstract}

\ifnature\else
\fi

\ifoptionfinal{}{\listoftodos}

\IEEEpeerreviewmaketitle



\newcommand{\tableCifarMods}{%
\raisebox{-1.25pt}{\Large\textcolor{exp-original}{$\bullet$}} Traditional ResNet-44                  & 92.2 & 0.0013 & 0.0014 \\
\midrule
\multicolumn{4}{p{\linewidth}}{\raggedright \hangindent=0.5cm \bfseries \boldmath Varying $\psi$ from \cref{eq:meth:lipschitz-loss} } \\
$\psi = 0.55, K=1$                                                                                   & 92.5 & 0.0022 & 0.0026 \\
$\psi = 4.0$                                                                                         & 92.4 & 0.0039 & 0.0048 \\
$\psi = 30.$                                                                                         & 90.2 & 0.0059 & 0.0080 \\
$\psi = 220$                                                                                         & 84.5 & 0.0083 & 0.0135 \\
\midrule
\multicolumn{4}{p{\linewidth}}{\raggedright \hangindent=0.5cm \bfseries \boldmath Varying $K$ from \cref{eq:meth:lipschitz-loss} } \\
$L_{2,adv}, \psi = 220, K=1$                                                                         & 84.5 & 0.0083 & 0.0135 \\
$K = 2$                                                                                              & 85.0 & 0.0084 & 0.0134 \\
$K = 4$                                                                                              & 85.0 & 0.0084 & 0.0135 \\
$K = 8$                                                                                              & 84.8 & 0.0082 & 0.0133 \\
\midrule
\multicolumn{4}{p{\linewidth}}{\raggedright \hangindent=0.5cm \bfseries \boldmath Regularization methods } \\
No additional regularization, $\psi = 30$                                                            & 90.2 & 0.0059 & 0.0080 \\
Stochastic depth \cite{Huang2016stochasticDepth}, $p_L = 0.8$                                        & 89.6 & 0.0059 & 0.0073 \\
Stochastic depth, $p_L = 0.5$                                                                        & 86.8 & 0.0051 & 0.0060 \\
ShakeDrop \cite{Yamada2018shakedrop}, $\alpha = 0, p_L = 0.8$                                     & 83.4 & 0.0043 & 0.0063 \\
\midrule
\multicolumn{4}{p{\linewidth}}{\raggedright \hangindent=0.5cm \bfseries \boldmath Varied network depth/width } \\
$\psi = 12,000$                                                                                      & 56.5 & 0.0110 & 0.0347 \\
ResNet-170                                                                                           & 55.2 & 0.0104 & 0.0332 \\
ResNet-44, double width                                                                              & 59.3 & 0.0118 & 0.0343 \\
\midrule
\multicolumn{4}{p{\linewidth}}{\raggedright \hangindent=0.5cm \bfseries \boldmath ``Dead zone'' from \cref{sec:meth:lipschitz} } \\
$\psi = 12,000, \sigma = 0$                                                                          & 56.5 & 0.0110 & 0.0347 \\
$\sigma = 0.01$                                                                                      & 66.8 & 0.0112 & 0.0288 \\
$\sigma = 0.05$                                                                                      & 79.4 & 0.0107 & 0.0197 \\
$\sigma = 0$, $\psi=220$ for similar accuracy                                                        & 77.1 & 0.0102 & 0.0194 \\
\midrule
\multicolumn{4}{p{\linewidth}}{\raggedright \hangindent=0.5cm \bfseries \boldmath Half-Huber ReLU from \cref{sec:meth:relu2b} } \\
$\psi = 12,000$, normal ReLU                                                                         & 56.5 & 0.0110 & 0.0347 \\
$\psi = 12,000$, HHReLU                                                                              & 78.0 & 0.0125 & 0.0261 \\
No HHReLU, but $\psi = 220$ for similar accuracy                                                     & 77.1 & 0.0102 & 0.0194 \\
\midrule
\multicolumn{4}{p{\linewidth}}{\raggedright \hangindent=0.5cm \bfseries \boldmath Varied $\zeta$ from \cref{sec:meth:lipschitz} } \\
$\psi = 12,000$, HHReLU, $\zeta = 0$                                                                 & 78.0 & 0.0125 & 0.0261 \\
$\zeta = 0.2$                                                                                        & 73.4 & 0.0128 & 0.0297 \\
$\zeta = 0.5$                                                                                        & 74.1 & 0.0118 & 0.0254 \\
$\zeta = 0.8$                                                                                        & 74.7 & 0.0107 & 0.0216 \\
$\zeta = 0.99$                                                                                       & 72.3 & 0.0083 & 0.0191 \\
\midrule
\multicolumn{4}{p{\linewidth}}{\raggedright \hangindent=0.5cm \bfseries \boldmath Output zeroing from \cref{sec:meth:outzero} } \\
Dead zone $\sigma = 1e-2$, $\zeta = 0$, $k_{out} = 0$, HHReLU, layer drop 0.8                        & 78.5 & 0.0123 & 0.0236 \\
Same, $k_{out} = 0.01$                                                                               & 78.9 & 0.0121 & 0.0222 \\
Double epochs (400 total)                                                                            & 80.8 & 0.0118 & 0.0214 \\
\midrule
\multicolumn{4}{p{\linewidth}}{\raggedright \hangindent=0.5cm \bfseries \boldmath Weight regularization from \cref{sec:meth:weightreg} } \\
Normal L2 weight regularization, HHReLU, $\zeta = 0.2$, dead zone $\sigma = 0.01$                    & 78.9 & 0.0131 & 0.0248 \\
Bias-only                                                                                            & 77.7 & 0.0126 & 0.0239 \\
\midrule
}

\newcommand{\tableCifarAdapt}{%
\raisebox{-1.25pt}{\Large\textcolor{exp-original}{$\bullet$}} Traditional ResNet-44                  & 92.2 & 0.0013 & 0.0014 \\
\multicolumn{4}{p{\linewidth}}{\raggedright \hangindent=0.5cm \bfseries \boldmath Others have HHReLU, normal weight regularization, no dead zone, $K=1$, $\zeta = 0$, $k_{out} = 0.01$, and use $L_{adv, z=2}$ unless otherwise specified. } \\
\midrule
\multicolumn{4}{p{\linewidth}}{\raggedright \hangindent=0.5cm \bfseries \boldmath Adaptive $\psi$ from \cref{sec:meth:adaptive} } \\
Fixed $\psi=12,000$, final training classification loss $1.007$                                      & 78.0 & 0.0125 & 0.0261 \\
$L_{target}=1.007$, initial $\psi=0.01$, final $\psi=14,000$                                         & 81.0 & 0.0124 & 0.0263 \\
Fixed $\psi = 14,000$                                                                                & 77.8 & 0.0131 & 0.0277 \\
\midrule
\multicolumn{4}{p{\linewidth}}{\raggedright \hangindent=0.5cm \bfseries \boldmath Varied network depth/width } \\
\multicolumn{4}{p{\linewidth}}{\raggedright \hangindent=0.5cm \bfseries \boldmath Fixed $\psi = 14,000$ } \\
ResNet-44                                                                                            & 77.8 & 0.0131 & 0.0277 \\
ResNet-170                                                                                           & 79.9 & 0.0132 & 0.0267 \\
ResNet-44, double width                                                                              & 81.9 & 0.0129 & 0.0257 \\
\multicolumn{4}{p{\linewidth}}{\raggedright \hangindent=0.5cm \bfseries \boldmath Adaptive $\psi$, $L_{target}=1.007$ } \\
ResNet-44                                                                                            & 81.0 & 0.0124 & 0.0263 \\
ResNet-44, $k_{\psi, 0} = 220, \epsilon_{worse}=0.01$                                                & 80.1 & 0.0127 & 0.0274 \\
ResNet-170                                                                                           & 82.1 & 0.0128 & 0.0277 \\
ResNet-44, double width                                                                              & 82.5 & 0.0129 & 0.0286 \\
\midrule
\multicolumn{4}{p{\linewidth}}{\raggedright \hangindent=0.5cm \bfseries \boldmath Different $L_{adv}$ from \cref{sec:meth:lipschitz} with $L_{target} = 1.007$ } \\
$L_{adv, z=1} = L_{adv, z=0, q=1}$                                                                   & 81.2 & 0.0069 & 0.0104 \\
$L_{adv, z=2}$                                                                                       & 81.0 & 0.0124 & 0.0263 \\
$L_{adv, z=3}$                                                                                       & 78.4 & 0.0139 & 0.0317 \\
$L_{adv, z=4}$                                                                                       & 77.9 & 0.0143 & 0.0326 \\
$L_{adv, z=5}$                                                                                       & 76.6 & 0.0144 & 0.0325 \\
$L_{adv, z=0, q=2}$                                                                                  & 81.1 & 0.0082 & 0.0176 \\
$L_{adv, z=1, q=1}$                                                                                  & 80.8 & 0.0080 & 0.0172 \\
$L_{adv, z=2, q=1}$                                                                                  & 79.9 & 0.0132 & 0.0295 \\
$L_{adv, z=2, q=2}$                                                                                  & 79.7 & 0.0132 & 0.0303 \\
\midrule
\multicolumn{4}{p{\linewidth}}{\raggedright \hangindent=0.5cm \bfseries \boldmath $L_{adv,tandem}$ from \cref{sec:meth:lipschitz} with $L_{target}=1.007$ } \\
ResNet-44, $\zeta = 0$                                                                               & 81.0 & 0.0124 & 0.0263 \\
ResNet-44, $\zeta = 0.2$                                                                             & 78.7 & 0.0127 & 0.0277 \\
ResNet-44, $\zeta = 0.2$, $L_{adv,tandem}$                                                           & 80.2 & 0.0131 & 0.0302 \\
ResNet-44, $\zeta = 0.2$, $L_{adv,tandem}$ with sum                                                  & 80.7 & 0.0121 & 0.0247 \\
\raisebox{-1.25pt}{\Large\textcolor{exp-best}{$\bullet$}} ResNet-44, $\zeta = 0.2$, $L_{adv,tandem}$, $L_{target}=1.5$ & 68.7 & 0.0151 & 0.0423 \\
\midrule
\multicolumn{4}{p{\linewidth}}{\raggedright \hangindent=0.5cm \bfseries \boldmath Adversarial / noisy training from \cref{sec:meth:adv-train} } \\
\multicolumn{4}{p{\linewidth}}{\raggedright \hangindent=0.5cm \bfseries \boldmath Madry \etal\ method, using $L_2$ adversarial training } \\
\raisebox{-1.25pt}{\Large\textcolor{exp-madry}{$\bullet$}} $L_2$, $\epsilon=0.01$                    & 87.4 & 0.0107 & 0.0153 \\
$L_2$, $\epsilon = 0.1$                                                                              & 88.6 & 0.0077 & 0.0205 \\
\multicolumn{4}{p{\linewidth}}{\raggedright \hangindent=0.5cm \bfseries \boldmath Madry \etal\ method, but with $L_{2,min}$ training } \\
$L_{2,min}$, $\epsilon = 0.01$                                                                       & 88.1 & 0.0092 & 0.0111 \\
$L_{2,min}$, $\epsilon = 0.1$                                                                        & 73.8 & 0.0121 & 0.0199 \\
HHAT, $L_{2,min}$, $\epsilon = 0.1$                                                                  & 84.4 & 0.0126 & 0.0179 \\
HHAT, $L_{2,min}$, $\epsilon = 0.1$, no HHReLU                                                       & 84.3 & 0.0122 & 0.0180 \\
$L_{2,min}$, $\epsilon = 0.25$                                                                       & 74.6 & 0.0151 & 0.0256 \\
HHAT, $L_{2,min}$, $\epsilon = 0.25$                                                                 & 83.1 & 0.0132 & 0.0204 \\
\multicolumn{4}{p{\linewidth}}{\raggedright \hangindent=0.5cm \bfseries \boldmath \Cref{eq:meth:lipschitz-loss} with adv. training, using $L_{target}=1.5$ and $L_{adv,tandem}$ } \\
$L_{2,min}$, $\epsilon = 0.1$                                                                        & 69.7 & 0.0195 & 0.0390 \\
\raisebox{-1.25pt}{\Large\textcolor{exp-madrymod}{$\bullet$}} HHAT, $L_{2,min}$, $\epsilon = 0.1$    & 68.4 & 0.0197 & 0.0450 \\
HHAT, $L_2$, $\epsilon = 0.1$                                                                        & 66.6 & 0.0164 & 0.0444 \\
HHAT, $L_{2,min}$, $\epsilon = 0.01$                                                                 & 67.5 & 0.0163 & 0.0443 \\
HHAT, $L_{2,min}$, $\epsilon = 0.1$, $L_{target} = 1.007$                                            & 79.0 & 0.0163 & 0.0306 \\
\multicolumn{4}{p{\linewidth}}{\raggedright \hangindent=0.5cm \bfseries \boldmath Gaussian noise, using $L_{target}=1.0$ and $L_{adv,tandem}$ } \\
Gaussian +- 0.05                                                                                     & 78.9 & 0.0135 & 0.0309 \\
Gaussian +- 0.25                                                                                     & 72.8 & 0.0129 & 0.0301 \\
\midrule
\multicolumn{4}{p{\linewidth}}{\raggedright \hangindent=0.5cm \bfseries \boldmath Combined adversarial training with output zeroing from \cref{sec:meth:outzero} } \\
\raisebox{-1.25pt}{\Large\textcolor{exp-madrymod}{$\bullet$}} HHAT with output zeroing               & 68.4 & 0.0197 & 0.0450 \\
HHAT without output zeroing                                                                          & 65.8 & 0.0197 & 0.0465 \\
\midrule
\multicolumn{4}{p{\linewidth}}{\raggedright \hangindent=0.5cm \bfseries \boldmath Active Learning from \cref{sec:meth:hitl} } \\
ResNet-44, double width, $L_{target} = 1.007$                                                        & 82.5 & 0.0129 & 0.0286 \\
Active learning version, \SI{448}{}/\SI{730}{} annotations                                                      & 82.7 & 0.0128 & 0.0285 \\
Active learning version, \SI{1331}{}/\SI{2731}{} annotations                                                    & 82.6 & 0.0133 & 0.0296 \\
}

\newcommand{\tableImagenet}{%
Standard ResNet-18                                                                                   & 65.6 & 0.0004 & 0.0013 \\
\midrule
\multicolumn{4}{p{\linewidth}}{\raggedright \hangindent=0.5cm \bfseries \boldmath Weight regularization from \cref{sec:meth:weightreg} } \\
$\psi = 0.7e6$, dead zone $0.002$, $k_{out} = 5e-5$$\dagger$                                         & 20.6 & 0.0013 & 0.0125 \\
Use 1e-6 instead of 1e-4 L2 regularization                                                           & 50.4 & 0.0039 & 0.0184 \\
\midrule
\multicolumn{4}{p{\linewidth}}{\raggedright \hangindent=0.5cm \bfseries \boldmath Automatic $\psi$ from \cref{sec:meth:adaptive} } \\
$\psi = 4.8e6$                                                                                       & 42.9 & 0.0041 & 0.0185 \\
$L_{target}=3.1$                                                                                     & 45.5 & 0.0041 & 0.0182 \\
$L_{target}=3.1$ with HHAT, $\epsilon=0.1$                                                           & 42.2 & 0.0053 & 0.0282 \\
$L_{target}=4.0$, no HHAT                                                                            & 35.5 & 0.0041 & 0.0243 \\
$L_{target}=5.0$                                                                                     & 23.7 & 0.0038 & 0.0388 \\
$L_{target}=5.0$, $L_{adv,z=2,q=1}$                                                                  & 22.0 & 0.0037 & 0.0538 \\
$L_{target}=5.0$, $L_{adv,z=5}$                                                                      & 19.7 & 0.0035 & 0.0526 \\
\midrule
\multicolumn{4}{p{\linewidth}}{\raggedright \hangindent=0.5cm \bfseries \boldmath COCO } \\
ResNet-44 from \cref{sec:meth:arch:coco}, baseline                                                   & 77.3 & 0.0003 & 0.0008 \\
With \cref{eq:meth:lipschitz-loss}                                                                   & 45.2 & 0.0029 & 0.0278 \\
AT only, $L_{2,min}$, $\epsilon=0.1$                                                                 & 60.8 & 0.0025 & 0.0089 \\
Combined \cref{eq:meth:lipschitz-loss} + AT                                                          & 45.2 & 0.0026 & 0.0228 \\
Combined \cref{eq:meth:lipschitz-loss} + HHAT                                                        & 45.2 & 0.0029 & 0.0250 \\
Balanced classes, \cref{eq:meth:lipschitz-loss} only                                                 & 25.8 & 0.0055 & 0.0335 \\
Balanced classes, \cref{eq:meth:lipschitz-loss} + HHAT                                               & 26.9 & 0.0054 & 0.0325 \\
}

\section{Motivation}\label{sec:intro}
Industry fields wanting to harness the explosion of {\em Machine Learning} (ML) techniques are concerned about the lack of accountability and explainability within the field \cite{Finlayson2019medicalAdversarial,Stilgoe2018machine}.  Biomedical papers report systems which surpass human experts, but have difficulty proving the added insight of their techniques beyond statistical correlations \cite{Finlayson2019medicalAdversarial,Tsao2018diabetic}.  This concern about explainability applies to a variety of ML algorithms, but we focus on the sub-field on {\em Neural Networks} (NNs).  State-of-the-art methods attempting to explain the reasoning behind NN decisions focus on the generation of heatmaps which indicate regions of input salient to the NN's output \cite{Selvaraju2017gradcam,Ribeiro2016lime}.  However, these heatmaps do not communicate information beyond a rough silhouette, making it difficult to infer much beyond the general region of an image considered.  These methods additionally rely on the linearization of a highly non-linear network, and capture relevant details only for the exact, corresponding input.  Minor perturbations can result in significant changes not only to the explanation, but also to the NN's output.

Adversarial attacks (or adversarial examples) are inputs to ML algorithms which are perceptually similar to examples that yield good performance from the algorithm, but produce drastically different output \cite{Szegedy2013adversarial,Goodfellow2014adversarial}.  These have been shown to exist on a variety of ML algorithms, and not only NNs \cite{Papernot2016transferability}.  The loophole of adversarial attacks poses a security risk at worst, and has left researchers scratching their heads at best.  Attempts to identify and remedy the problem of adversarial examples generally agree on the existence of manifolds shared by the dataset which are incongruous to perceptual manifolds \cite{Athalye2018obfuscatedGradients,Khoury2018geometry}.  Work on adversarial examples has not generally focused on creating situations in which adversarial examples are perceptually similar to the targeted class; only Tsipras \etal\ \cite{Tsipras2018robustness}\ touched on this as a curiosity associated with adversarial training.


\fig*[label=fig:research:adv-motivation]
  {\graphic{figs/explain-vs-gradcam-ilsvrc.pdf}}
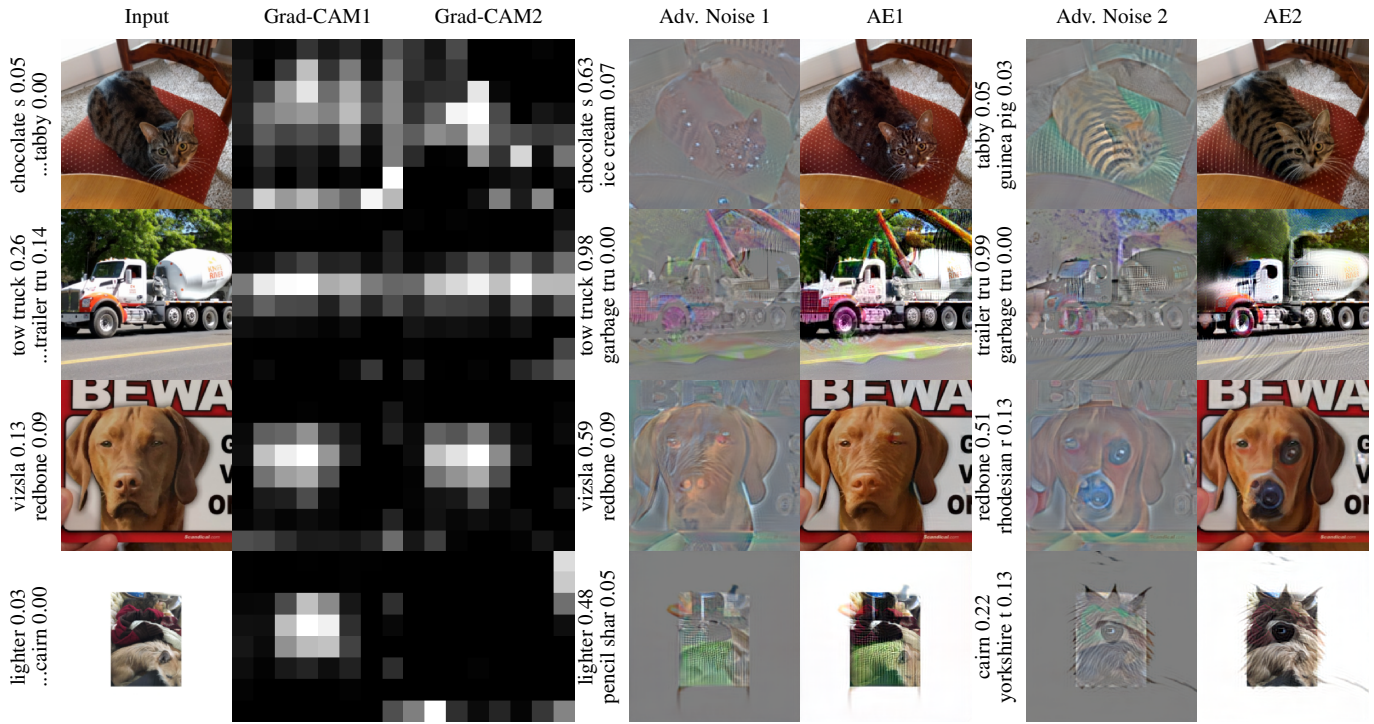
  {Comparing the explanatory power of Grad-CAM \cite{Selvaraju2017gradcam}\ and AE when applied to a robust NN on the ILSVRC classification task.  For each input image, a class prediction was run.  The highest prediction is shown on the left of each row, followed by either the second highest prediction or the true class.  The ``Input'' column shows the input image.  The ``Grad-CAM1'' column shows a Grad-CAM explanation for positive regions of the first predicted class.  The ``Grad-CAM2'' column shows a Grad-CAM explanation for positive regions of the second-most-likely (or true) class prediction.  Following that are AEs for the same two class predictions, created using $g_{explain+}$ from \cref{sec:meth:adv:explain} Each adversarial perturbation shows the new top-two class predictions for the modified image, the noise perturbation from the original image, and the resulting modified image (the AE).  Note that the adversarial perturbations often do not align with the Grad-CAM-highlighted saliency regions, e.g. in row 4, a picture of a Cairn Terrier.  This happened because Grad-CAM is a linearization of the non-linear NN.  In contrast, the AEs worked with the non-linear perturbations to the images, and revealed the textures and structures which were important for identifying each predicted class.  Row 1 demonstrates that a ``Tabby Cat'' was not predicted as the stripes and whiskers on this animal were too low contrast.  Row 2 demonstrates that the NN did not predict a ``Trailer Truck'' because there was no clear separation between the cab and the trailer, a quality which was associated with ``Tow Truck'' during training.  Row 3 was correctly predicted as a ``Vizsla,'' with a lack of the distinct colorization preventing higher confidence.  The colorization also matched that of the ``Redbone'' dog, which would require darker nose and eye areas.  Row 4, a highly-cropped ``Cairn Terrier,'' looked more like a lighter due to the patterns of the clothing behind the dog.  To be identified as a ``Cairn Terrier,'' the NN would have needed an image which captured more pronounced texturing of the animal's fur - a partial indication that the image would have needed to be zoomed in on the main subject.  An extended discussion of a figure like this one and the insights of AEs may be found in \cref{sec:results:cifar-explain}.}

We contribute a set of novel techniques which allow for {\em Adversarial Explanations} (AEs) to illustrate key salient features for classification, a much more reliable method of explaining an NN's decision.  Unlike previous state-of-the-art techniques, AEs work with network non-linearities to represent the NN's decision surface with greater fidelity than heatmaps can provide, as shown in \cref{fig:research:adv-motivation,fig:research:adv-overview}.  Further discussion of previous state-of-the-art explanations is presented in \cref{sec:related:expl}.
In addition to producing visually rich explanations, our techniques surpass the state of the art in terms of classification performance in the presence of adversarial examples.
We propose and demonstrate classification networks for the ILSVRC 2012 challenge with \valImagenetOurBenefitAuc\ improved robustness to adversarial attacks compared to the state of the art, as shown in \cref{fig:research:acc-vs-rmse} and discussed in \cref{sec:related:adv}.  Both AEs and improved robustness were achieved via the methods described in \cref{sec:meth}.
We explore the explanatory power and trade-offs of the proposed techniques, including the ability to train networks to be either more accurate or robust to attacks, in \cref{sec:results}.  Due to the visual quality of AEs, they may also be used to synthesize new examples for an active learning pipeline to improve a classifier's robustness, which we demonstrate in \cref{sec:meth:hitl,sec:results:cifar:hitl}.  Together, the methodology outlined in this work demonstrates the viability of producing cogent explanations via adversarial attacks on robust networks.

%

\fig[label=fig:research:adv-overview]
    {\graphic[width=.99\linewidth]{figs/explain-ours-overview.svg}}
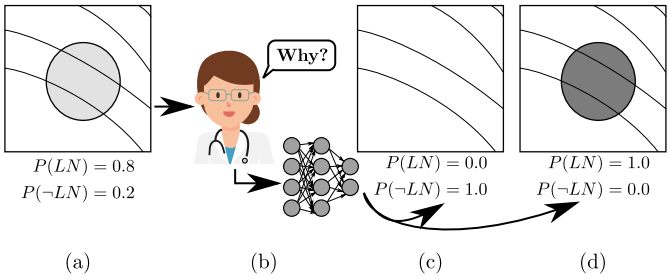
    {While adversarial examples are considered a nuisance by most, they have the potential to provide reliable explanations with the same richness of information as the original input.  For example, when an NN trained at finding lung nodules in radiographs needs investigation (a, b), an attack may be targeted at a desired new network output---such as changing a nodule classification to a non-nodule classification (c) or emphasizing the nodule (d)---to produce a new image which is minimally changed but produces the desired output.  By comparing these inputs and looking at the differences, a human operator can identify relevant features in the input with greater fidelity than prior methods of explanation.}

\fig[label=fig:research:acc-vs-rmse]
    {%
        \subfig[label=fig:research:acc-vs-rmse:imagenet]{\graphic{plots/comparison-imagenet.pgf}}
        \subfig[label=fig:research:acc-vs-rmse:cifar]{\graphic{plots/comparison-cifar.pgf}}}
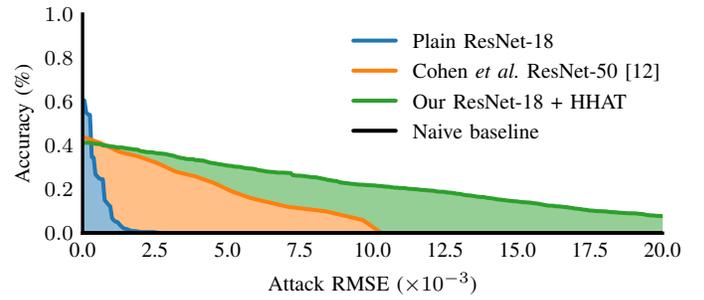
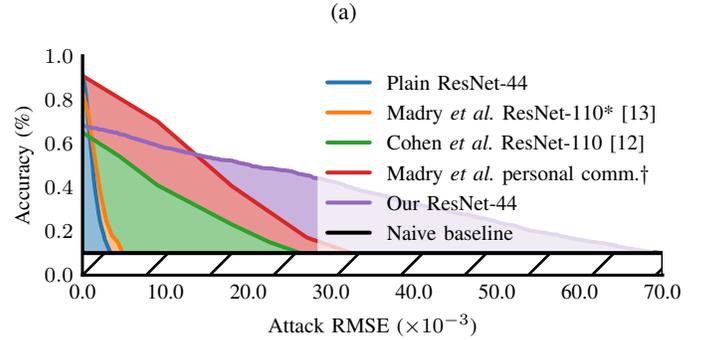
    {Compared to the state of the art, our method achieved NNs which were tolerant of greater adversarial perturbations on both ImageNet (\cref{fig:research:acc-vs-rmse:imagenet}) and CIFAR-10 (\cref{fig:research:acc-vs-rmse:cifar}).  By measuring the ARA, a model's resistance to adversarial attacks is taken into consideration alongside its ability to make predictions better than the naive baseline (the hatched region).  See \cref{sec:related:adv} for additional details. \\
    * Madry \etal\ used a network that was $10\times$ as wide as a traditional ResNet-110, and also trained against an $L_{\inf}$ adversary rather than an $L_2$ adversary \cite{Madry2017towardsResistantAdv}. \\
    $\dagger$ This curve came from personal communication with A. Madry on a standard ResNet-50; see \cref{sec:related:adv} for details.}

\section{Related Work}

Two branches of ML inquiry led to this work: adversarial attacks and explanation methods.  To better demonstrate the flaws in existing explanation techniques, we cover adversarial attacks before competing explanation methods.  A mathematical branch, Lipschitz continuity, was also important in developing this work.  In exploring the merits of better NN explanations, active learning methods were also considered.  These four related areas are discussed in \cref{sec:related:adv,sec:related:expl,sec:related:hitl,sec:related:lipschitz}.

\subsection{Adversarial Attacks}\label{sec:related:adv}

Adversarial attacks, or adversarial examples, were first documented by Szegedy \etal\ \cite{Szegedy2013adversarial}, who showed that an NN's output may be arbitrarily changed through imperceptible changes to the input.  Initial criticism that these digitally-induced deviations might be a pathological problem were put to rest by a group from LabSix in 2018 \cite{Athalye2017synthesizingAdversarial}.  The LabSix group fabricated a real-world object which was adversarially misclassified at a variety of angles and scales, demonstrating that the problem of adversarial examples had real-world consequences and deserved further study.  Many reports have posited that adversarial examples are a natural extension of the internal flexibility of NNs \cite{Szegedy2013adversarial,Goodfellow2014adversarial,Carlini2016robustness,Athalye2017synthesizingAdversarial,Tsipras2018robustness}, and a survey of the topic of adversarial attacks and defenses covers many related topics \cite{Chakraborty2018adversarialSurvey}.  These reports all support that NNs solve an underconstrained problem: many possible solutions to the training data exist on manifolds which are distorted toward imperceptibility in standard visual space.  Adversarial attacks exploit the incongruence between these learned spaces and the visual space containing the NN's input.

The struggle between adversarial attacks and methods of resisting them is perhaps best illustrated by the saddle point formulation proposed by Madry \etal\ \cite{Madry2017towardsResistantAdv} (identical to their approach in \cite{Tsipras2018robustness}, though that work contains further analysis; notation adapted to be consistent throughout the current paper):

\begin{align}
    \min \mathbb{E}_{(x, t)\sim D}\left[ \max_{\delta\in S}L(\theta, x + \delta, t)\right]. \label{eq:madry}
\end{align}

\noindent That is, find the network parameterized as $\theta$ which, according to dataset $D$, produces the best approximation of some target $t$ when the worst-case noise $\delta$ constrained by an allowed attack space $S$ is added to an input $x$.  This formulation illustrates the difficulty of working against a high-quality adversary, which is relatively unrestricted in its exploitation of the network's properties around $x$.

Methods of generating adversarial attacks approximate the inner maximization problem from \cref{eq:madry}.  In this work, we focused on white-box attacks, which are attacks where the attacker has full knowledge of the model's internal parameters and configuration.  These were chosen specifically because they are the most difficult to defend against.  Carlini \etal\ \cite{Carlini2016robustness}\ compared several different methods of generating attacks, including Goodfellow \etal's {\em Fast Gradient Sign Method} (FGSM) \cite{Goodfellow2014adversarial} and their own {\em Projected Gradient Descent} (PGD) \cite{Carlini2016robustness}.  Carlini \etal\ \cite{Carlini2016robustness}\ showed that attacks were transferable between networks, regardless of network architecture.  One interpretation of this would be that NNs emphasize high-frequency signals within the data over low-frequency signals, biasing them toward changes which are imperceptible in the domain of the input.  Recent work by Tsipras \etal\ \cite{Tsipras2018robustness}\ argued that this might be due to the natural tendency of high-accuracy classifiers to exploit small differences as a means of greedily leveraging available information.
Stutz \etal\ \cite{Stutz2019disentangling} interestingly studied the creation of ``on-manifold'' adversarial examples, which conform to the original input distribution as defined by a {\em Variational Autoencoder} (VAE) - {\em Generative Adversarial Network} (GAN) hybrid.  Unlike with off-manifold, or traditional, adversarial examples, they found that generalization accuracy would actually be increased by training with on-manifold adversarial examples \cite{Stutz2019disentangling}.  However, for the purposes of the current work, even ``off-manifold'' adversarial examples still fit into the NN's valid input space, and an attacker often creates attacks outside of the original data manifold to exploit this incongruence.

Methods of providing robustness against adversarial attacks approximate the outer minimization problem from \cref{eq:madry}.  To our knowledge, state-of-the-art methods of resisting adversarial attacks currently revolve around either adversarial training \cite{Madry2017towardsResistantAdv,Tsipras2018robustness,Pei2017deepxplore} or randomized smoothing \cite{Cohen2019certified}.

Madry \etal\ \cite{Madry2017towardsResistantAdv}\ investigated using both the FGSM and PGD methods of generating adversarial examples, and the effects of using these methods to train networks, a technique called adversarial training.  Note that, under adversarial training, a network is consistently trained based on its worst performance point in the neighborhood of each input.  Madry \etal\ showed that adversarial training could reduce the transferability of attacks between different networks, but only slightly \cite{Madry2017towardsResistantAdv}.  They also argued that any defense mechanism shown to be robust against PGD would be robust against other first-order attacks \cite{Madry2017towardsResistantAdv}.  That group later expanded their theories on adversarial training in work by Tsipras \etal\ \cite{Tsipras2018robustness}, demonstrating salient features materializing in adversarial examples with large limits on allowed attacks.  These attacks used perturbation magnitudes which greatly surpassed the threshold at which the classifier's accuracy would necessarily change, but for the first time demonstrated that classifiers could potentially be used to alter input images, adding or subtracting salient features.

Cohen \etal\ \cite{Cohen2019certified}\ recently improved on a body of work called randomized smoothing, a provable method of inducing $L_2$ robustness based on evaluating a smoothed version of a network trained with Gaussian noise.  Unlike techniques such as Madry \etal's \cite{Madry2017towardsResistantAdv}\ or our work in \cref{sec:meth}, Cohen \etal's method allows for a certifiable calculation of an adversarial resistance bound.  That is, there may or may not be attacks against networks that exist, but are difficult to find with PGD.  If these attacks exist, Cohen \etal's method provides a high level of confidence that the smoothed network would also protect against them, even though these attacks cannot currently be generated.  The price of this certainty comes with somewhat inflated processing time: predicting requires about 100 evaluations for each input to properly compute the smoothing function \cite{Cohen2019certified}.

Other defensive techniques have been proposed but were either inadequately tested or shown to be broken.  An approach which denoised inputs won the NeurIPS 2017 adversarial robustness challenge \cite{Liao2017denoiser,Kurakin2018nipsRoundup}, which was successful but defended against a static set of attacks targeted at a standard network rather than against attacks specific to the defended network.  A number of stochastic and non-differentiable defenses have been proposed and subsequently shown to be vulnerable to attacks which take these qualities into account \cite{Athalye2018obfuscatedGradients}.  Still other defense papers have focused on defense against specifically single-step attacks \cite{Tramer2018ensemble}, were marginally less effective versions of the previously mentioned, state-of-the-art approaches \cite{Wong2018scalingProvable}, or focused on the natural defensive qualities of different architectures rather than ways of improving their defenses \cite{Su2018robustness}.

We note that a roundup of best-practices for ensuring that new defenses are effective was recently authored by Carlini \etal\ \cite{Carlini2019evaluatingRobustness}.  In the context of the current work, we've complied with many of their recommendations, excepting non-gradient based attacks and an investigation of attack transferability.  As our proposed techniques only affect network regularization (\cref{sec:meth}) or make gradients less obfuscated (\cref{sec:meth:relu2b}), sticking to a gradient-based PGD-variant attack seemed sufficient.  Attack transferability was not investigated as the proposed models have identical architectures and processing, and therefore, for a given input, increasing the required attack perturbation magnitude necessitates that an attack would not transfer.

For comparing adversarial defense techniques, the current work used accuracy-versus-attack-magnitude plots, such as \cref{fig:research:acc-vs-rmse}.  This type of plot shows how an individual classifier's accuracy would fall as the allowable attack space, $S$ from \cref{eq:madry}, is increased.  For consistency across datasets, regardless of their input dimensions, we used the {\em Root-Mean-Squared Error} (RMSE) for the shown attack distances, which is equal to the $L_2$ norm of the perturbation divided by the square root of the number of elements in the perturbation.  The RMSE is a more natural choice as it is scaled such that an RMSE of $0$ means no change and an RMSE of $1$ means the change between an all-black and all-white image, regardless of size.  Figures shown in this work have scaled e.g. $\epsilon$ and other values reported in other works to the RMSE scale.  Cohen \etal\ \cite{Cohen2019certified} presented this plot for $L_2$ attacks against both ImageNet and CIFAR-10 results, and Madry \etal\ \cite{Madry2017towardsResistantAdv} presented this plot for $L_2$ attacks against a CIFAR-10 classifier with $10\times$ the normal number of filters and trained against $L_{\inf}$ adversarial examples.  Personal communication with A. Madry yielded the additional curve on \cref{fig:research:acc-vs-rmse}, which was for a standard CIFAR-10 ResNet-50 trained against $L_2$ attacks with $\epsilon=0.009$; we independently trained a similar network using their methods and achieved similar results, and show improved results for adversarial training with a slightly different adversary (\cref{sec:results:cifar:adv-train}).

To compare these curves using a single number, we've used the area between the curves of a naive classifier and the classifier in question, a metric we've termed the classifier's {\em Accuracy-Robustness Area} (ARA).  A larger ARA value is desirable.  The ARA is illustrated in \cref{fig:research:acc-vs-rmse:cifar}, where the shaded area for each classifier is the area computed for the ARA.  Intuitively, the ARA measures a combination of the classifier's predictive power and its ability to overcome an adversary.  Importantly, when constrasted against existing robustness metrics, the ARA takes into account the classifier's performance against all adversarial examples, without bounding them by some arbitrary $\epsilon$.  For triangular shapes such as those in \cref{fig:research:acc-vs-rmse}, where the accuracy smoothly declines from the classifier's accuracy on clean data to the naive classification baseline at some point, the ARA is about equal to $\frac{1}{2}S(A - N)$, or half the product of $S$, the adversarial perturbation magnitude at which the classifier has no predictive power, and the difference between $A$, the classifier's clean accuracy, and $N$, a naive classifier's accuracy on the problem.  Thus, when clean accuracies are identical, a network with an ARA $3\times$ larger than another network's ARA indicates that the first network retains predictive power against adversaries which produce $3\times$ more noise than where the second network would fail.  For a more realistic example, consider \cref{fig:research:acc-vs-rmse:cifar}.  The curve ``Madry \etal\ personal comm.'' has a clean accuracy of \SI{90.9}{\%} and loses its predictive powers against an adversary with $S \approx 0.03$, and an ARA calculation yields \SI{0.0124}{} (note that $\frac{1}{2}(0.03)(0.909 - 0.1) = 0.0121$).  The curve ``Our ResNet-44'' has a clean accuracy of \SI{68.4}{\%} and loses its predictive powers at $S \approx 0.07$, and an ARA calculation yields \SI{0.0197}{} (here $\frac{1}{2}(0.07)(0.684 - 0.1) = 0.0204$).  The crossover point -- at which our ResNet-44 becomes more accurate than the adversarially-trained classifier from Madry \etal\ -- occurs when attack magnitudes exceed $S \approx 0.013$.  See \cref{app:cifar-4} for an example of different perturbation magnitudes; generally, an RMSE of $0.013$ would be indistinguishable from the original input.

We found that, on CIFAR-10, a standard ResNet-44 had an ARA of \valCifarOurAucTraditional, extrapolating numbers from Madry \etal's best $L_2$ resistant network (from personal communication) yielded the aforementioned ARA of \valCifarMadryAuc, and Cohen \etal's numbers yield an ARA of \valCifarCohenAuc.  A reproduction of Madry \etal's best $L_2$ resistant network, but as a ResNet-44 instead of ResNet-50 and using our algorithm for evaluating ARA in \cref{sec:meth:adv:accuracy}, yielded an ARA of \valCifarMadryReproAuc.  On ImageNet 2012, a standard ResNet-18 had an ARA of \valImagenetOurAucTraditional\ and Cohen \etal's method resulted in an ARA of \valImagenetCohenAuc.  We note that, to the best of our knowledge, Madry \etal's \cite{Madry2017towardsResistantAdv,Tsipras2018robustness} group has not yet produced a robust network on the full ImageNet challenge.



\subsection{Explanation Methods}\label{sec:related:expl}

Inadequate understanding of the internal operation of NNs, or the larger toolbox of ML solutions in general, has recently come under focus as a primary difficulty of using them \cite{Ribeiro2016lime,Selvaraju2017gradcam,Simonyan2013deep,Landecker2014interpretable,Bau2017networkDissection,Murdoch2019interpretableSurvey}.
Preliminary attempts at addressing this problem involved looking at saliency maps computed via backpropagation to see which input pixels had the largest effect on the classification \cite{Simonyan2013deep,Hong2015tracking} or looking at the receptive fields to which internal nodes respond \cite{Zeiler2014visualizingNnRfs,Olshausen1996emergence,Luo2016understandingRfs,Bau2017networkDissection}.  However, these techniques often produce very noisy images that are difficult to interpret, and have been shown to be fragile explanations in the presence of adversarial examples \cite{Ghorbani2019interpretation}.

Works such as Ribeiro \etal's {\em Local Interpretable Model-Agnostic Explanations} (LIME) \cite{Ribeiro2016lime} or Selvaraju \etal's {\em Gradient-weighted Class Activation Mapping} (Grad-CAM) \cite{Selvaraju2017gradcam} proposed improvements over raw saliency maps.  LIME is a generalized method suitable for both image and non-image inputs that boils down to set theory: if part of the input were masked, would the overall classification get better or worse?  By noting which parts of the input make the most significant difference, the LIME algorithm derives a linear classifier which approximates the non-linear NN, and the linear, approximating classifier is then used to produce a mask for the input that highlights salient regions \cite{Ribeiro2016lime}.  On the other hand, the Grad-CAM algorithm harnesses backpropagation directly to derive an expression for localizing the most salient regions \cite{Selvaraju2017gradcam}.  Its innovation came from measuring image region contribution at a layer closer to the classifying end of the NN than the input.  Both of these ultimately used linearization techniques in an attempt to describe the non-linear NN's behavior.

These algorithms produced reasonable explanations for the examples provided in their papers \cite{Ribeiro2016lime,Selvaraju2017gradcam}.  The paper proposing LIME additionally presented a convincing argument that accuracy alone may not be representative of a classifier's quality, and explanations can highlight generalization errors caused by artifacts within the original dataset \cite{Ribeiro2016lime}.  However, neither the LIME nor Grad-CAM algorithms account for non-linear network behaviors, and the corresponding papers did not speak to their validity outside of the exact input being evaluated.  To further test the validity of Grad-CAM and LIME for explaining NNs, which are highly non-linear by design, we considered the relation between explanations and adversarial examples: if a method explaining an NN's decision were reliable, then an adversarial example should primarily change the regions of the image marked salient by the explainer.

This theory was tested on the lung nodule dataset published by the {\em Japanese Society of Radiological Technology} (JSRT).  This dataset consists of 247 chest X-rays,  154 of which have a single, annotated nodule, each of which show up as dark, solitary shadows.  Chest X-rays are notoriously difficult to read, and so the JSRT dataset is evaluated by nominating 5 candidate points which might be proximal to a tumor.  An algorithm's output is considered correct if a tumor center is within \SI{2.5}{\centi\meter}, or \SI{143}{px}, of any such candidate point.  Scores are thus presented as sensitivity given at most 5 false positives per X-ray.
We trained a network on this problem which scored 66\% sensitivity by this rubric and applied LIME and Grad-CAM to the NN, yielding \cref{fig:related:lime} and \cref{fig:related:gradcam} respectively.

Considering the output of LIME for this classifier, \cref{fig:related:lime}, we see that the produced explanation neither makes intuitive sense nor instills confidence in the classifier.  For the same network, Grad-CAM produces a smoother, more intuitive explanation (\cref{fig:related:gradcam}).  One could reasonably infer from this explanation that the network assigned saliency to the high-contrast borders of the nodule, a reasonable approach for the problem.  However, this human-oriented interpretation of Grad-CAM's output seems unrepresentative of the network's actual operation: when compared with the adversarial example generated for the same NN in \cref{fig:related:adv}, there is no correlation in overlap between the adversarial perturbations and the salient region demarked by Grad-CAM.  As such, we posit that the explanations yielded by LIME and Grad-CAM do not reliably represent dominant factors contributing to the NN's output.  While LIME and Grad-CAM have sound theory, they both rely on linearizations of a highly non-linear network, and an imperceptible change in the input invalidates the entire explanation from either algorithm.

\fig[label=fig:related:lime]
    {\graphic[]{figs/explain-lime.pdf}}
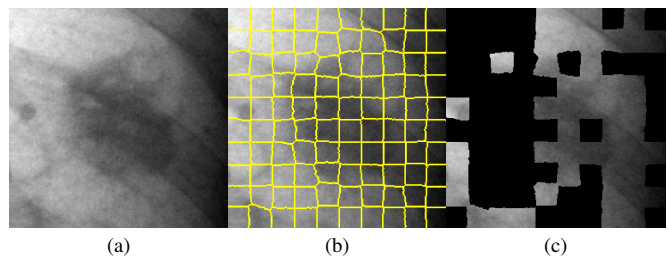
    {A demonstration of LIME on an NN diagnosing lung nodules in the JSRT dataset.  LIME's algorithm consists of taking an input (a), dividing it up into super-pixels (b), and then using linear approximations to determine which subset of super-pixels most significantly affects the network's classification (c).  How well does this explain the decision?}
\fig[label=fig:related:gradcam]
    {\graphic[]{figs/explain-gradcam.pdf}}
    {Selvaraju \etal's Grad-CAM \cite{Selvaraju2017gradcam} processes an input (a), uses back-propagation to produce localized gradient information (b), and presents that information as a heat map of salient regions (c).  How well does this explain the decision?}
\fig[label=fig:related:adv]
    {\graphic[]{figs/explain-adversarial.pdf}}
    {An adversarial example corresponding to the same network and input as \cref{fig:related:lime,fig:related:gradcam}.  Here, the resulting adversarial input (a) is the sum of the original input (b) and a crafted noise term scaled by a small coefficient (c).  The result is an imperceptible change in input leading to a completely different NN output.  Also worth noting is that the location of the noise in (c) does not correlate well with either the LIME or Grad-CAM explanations.}

Attention models involve modifying the NN's architecture to both enhance overall NN performance and provide an attention mask which may be viewed as an explanation mechanism \cite{Vaswani2017attention,Cui2016attention}.  Where saliency maps demonstrate the most-used pixels, attention models use a gating technique which involves multiplying the inputs by a learned mask to shape how inputs are forwarded to the classifying portion of the network.  Viewing the mask of the attention model shows which inputs were weighted more heavily for a given input.  However, attention models are still vulnerable to adversarial attacks which exploit only a small portion of the masked region.  Jetley \etal\ \cite{Jetley2018learnAttention} analyzed their attention masks in relation to FGSM adversarial attacks, and found only very marginal benefits.

Also noteworthy is Bau \etal's ``Network Dissection'' work \cite{Bau2017networkDissection}.  By cross-referencing the activation map of convolutional neurons in an NN with object annotations, an IoU was computed on a per-neuron basis.  They reasoned that neurons relating to objects via a large IoU were responsible for detecting that type of object or texture.  However, the reported IoUs are quite small, with the majority of reported values being below 0.2---higher than coincidence, but lower than an authoritative explanation would merit.  Bau \etal\ also made no mention of how adversarial examples relate to their work.  The approach is worth continued research, but is not yet an end-all means of explaining NNs.  More recently, Bau \etal\ have applied their dissection methods to GANs, and demonstrated that omitting neurons with high IoUs in GANs can predictably modify the generated images \cite{Bau2018ganDissection}, lending evidence that per-neuron explanations might also be feasible in a correctly-formulated classification network.


\subsection{Lipschitz Continuity}\label{sec:related:lipschitz}

Briefly, Lipschitz continuity is the bounding of a function's value, such that the function's value is not allowed to change between two points more than a constant value times the distance between those points.  This is often approximated as a global bounding of the derivative.

Prior work has combined Lipschitz continuity with NNs.
Weng \etal\ \cite{Weng2018Lipschitz} proposed analyzing the stability of NNs using a metric derived from gradient measurements at different data points, which they claimed was analogous to a Lipschitz constraint.  However, their method comprised of sampling the gradient at different data points, a linearization which provided no guarantees about behavior between those points, a flaw similar to those we mentioned regarding LIME and Grad-CAM.  They also never attempted to control the gradients or assess a causal relationship between the gradient magnitude and a network's robustness, instead focusing on a correlative argument.
Cisse \etal\ \cite{Cisse2017parseval} implemented their Parseval networks by restraining the Lipschitz constant of each individual, weighted layer to be less than $1$.  In that work, it was found that adding the Lipschitz constraint made NNs negligibly more resistant to attacks, with the Parseval network's accuracy falling at much the same rate as a vanilla network's as attacks of larger magnitude were allowed \cite{Cisse2017parseval}.
Behrmann \etal\ \cite{Behrmann2019invertible} proposed that bounding the per-layer Lipschitz constraint of an NN to $1$ resulted in beneficial invertibility properties, but did not analyze the effect of that bound on the network's adversarial robustness.

We differ from prior work in this area as we consider the entirety of the NN as subject to end-to-end Lipschitz constraints, rather than each layer individually.  We propose a novel Lipschitz constraint form which, rather than assigning an upper limit, aggressively minimizes the Lipschitz constraint.  We demonstrate this technique as providing significant resistance against adversarial examples, using a metric which accounts for non-linearities in gradient behavior.  We also propose several other modifications to NNs, which complement the practical minimization of a Lipschitz bound for NNs.

\subsection{Active Learning}\label{sec:related:hitl}

While we have demonstrated that heat maps are poor indicators of a robust explanation, they have been proven useful for active learning.  Li \etal's ``Tell Me Where To Look'' \cite{Li2018tmwtl} demonstrates the merit of using annotated salient regions---a straightforward method of providing active learning functionality---as part of weakly-supervised training for NNs.  They introduced an algorithm which extends Grad-CAM's heat map to be differentiable, and then used gradient descent so that the highlighted region approaches that of annotated segmentations on a classification dataset. This method could be considered a proof-of-concept for active learning applications based on NN explanations, where each new human annotation ultimately becomes part of the training set for the algorithm.  For segmentation, their algorithm resulted in an impressive leap from a mean {\em Intersection-over-Union} (IoU) of 0.555 with prior methods to 0.621 with theirs \cite{Li2018tmwtl}.  Additionally, for \SI{10000}{} classification training examples, they only used segmentation masks for \SI{1464}{} of the examples \cite{Li2018tmwtl}, illustrating that even partial annotations provided some benefit.

Ribeiro \etal\ \cite{Ribeiro2016lime}\ used LIME in an active learning context by providing operators, who were unfamiliar with machine learning, explanations on factors salient to a classifier's decision.  They first created classifiers that identified whether the topic from a text document was ``Atheism'' or ``Christianity,'' from a dataset of 20 newsgroups.  Operators on Amazon Mechanical Turk were then shown explanations created via LIME, and marked the words which LIME determined were salient but which should not have been relevant to the task.  New classifiers were then trained on a modified version of the dataset, which did not include the words deemed irrelevant by operators.  This process was repeated several times, with an average of \SI{200}{} words being removed between the original dataset and the final classifier; Ribeiro \etal\ did not mention the number of words in the original dataset.  The real world accuracy of the classifiers improved from a baseline of approximately \SI{57}{\%} to approximately \SI{70}{\%} when the instances being explained were chosen randomly; they also proposed an instance-picking algorithm which boosted this performance further to \SI{73}{\%}.  Importantly, this experiment demonstrated that active learning processes could be used to significantly improve classifiers after few iterations.

We explored an alternative active learning pipeline to improve the adversarial resistance of an NN, instead of improving accuracy.  As there is no image-oriented analog for removing entire words from a dataset, we instead focused on introducing new training data.  As far as we are aware, we are the first to generate new training data from existing training data as part of an active learning pipeline.  While the adversarial training of Madry \etal\ \cite{Madry2017towardsResistantAdv}, or even standard data augmentations, could be viewed as introducing new data, both of those are implemented with perturbations designed not to change the underlying true class of data points.  In contrast, we relied on AEs with magnitudes sufficient to change the underlying class, necessitating human input.  Our active learning process is outlined in \cref{sec:meth:hitl}.

\section{Methods}\label{sec:meth}

Adversarial examples in state-of-the-art networks, as in \cref{fig:related:adv}, do little to explain the inner workings of the NN for which they are generated.  However, the potential exists for adversarial examples to be a very powerful, non-linear method of explanation.  Tsipras \etal\ \cite{Tsipras2018robustness}\ demonstrated that non-minimal adversarial examples contained salient features when networks were adversarially trained.  Here, non-minimal means adversarial examples which have not been optimized for minimal perturbation distance, but only for maximal loss on the objective function.  The current work considers whether decision boundaries may be pushed out even further, such that minimal adversarial examples at class boundaries might demonstrate the removal of features salient to the original classification.  While previous methods of explaining NNs rely on linearization techniques, adversarial examples make full use of the NN's non-linearities.  With targeted attacks, the boundary located could signify different salient aspects of the input stimulus, as in \cref{fig:research:adv-motivation,fig:research:adv-overview}, if the decision manifold of the network were sufficiently congruent with the visual manifold.  A practical demonstration of this theory may be seen through examples on the JSRT dataset in \cref{fig:research:ours-less,fig:research:ours-more}; these were implemented using the techniques described throughout this section.


We will first discuss our method of generating and evaluating both adversarial attacks and explanations in \cref{sec:meth:adv}.  Following that, the methods used to train robust NNs may be found in \cref{sec:meth:lipschitz,sec:meth:weightreg,sec:meth:relu2b,sec:meth:outzero,sec:meth:adaptive,sec:meth:adv-train,sec:meth:hitl}, and finally a discussion of the datasets used in this work and the NN architectures chosen for those datasets is found in \cref{sec:meth:arch}.

\fig[label=fig:research:ours-less]
    {\graphic[]{figs/explain-ours-less.pdf}}
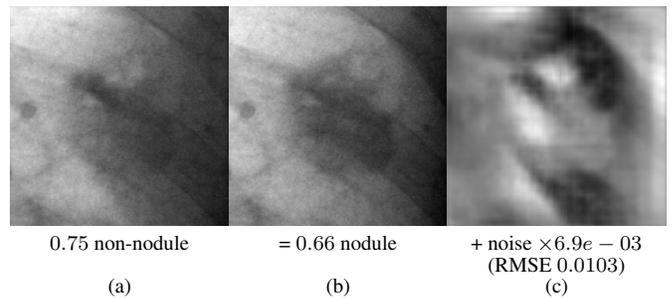
    {In contrast to \cref{fig:related:adv}, our preliminary technique of increasing adversarial distance concentrates the perturbations needed to change the network's output.  The RMSE (c) between the original input (a) and the adversarial attack (b) is also much greater for a smaller change in network output.}
\fig[label=fig:research:ours-more]
    {\graphic[]{figs/explain-ours-more.pdf}}
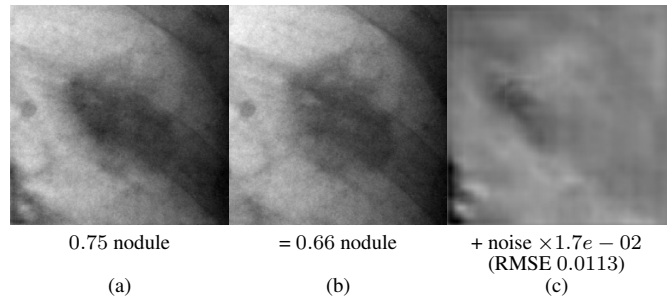
    {Example of emphasizing a lung nodule via the same method of adversarial attack.  Same network as \cref{fig:research:ours-less}.}


\subsection{Adversarial Attack Generation and Evaluation}\label{sec:meth:adv}

Adversarial attacks were conducted with two separate goals within this paper: \cref{sec:meth:adv:accuracy} contains the methodology for adversarial attacks aimed at reducing the classification accuracy of a network, and \cref{sec:meth:adv:explain} contains the methodology for adversarial attacks aimed at producing classification explanations.

\subsubsection{Adversarial Attacks on Accuracy}\label{sec:meth:adv:accuracy}

Untargeted adversarial attack generation for the evaluation of models followed \cref{alg:meth:adv:all}; this was a variant of Carlini \etal\ \cite{Carlini2016robustness}, and also leveraged normalizing gradient steps by their magnitude, first proposed by Rony \etal\ \cite{Rony2019decouplingAdvDirection} in the context of adversarial attacks.  Rather than pursuing both target loss maximization and $L_2$ error minimization simultaneously, we found that alternating between these two to traverse some restriction on the adversarial example's network output allowed for better automatic balancing between the two errors, resulting in smaller perturbation magnitudes.  In contrast to the attacks presented by Carlini \etal\ \cite{Carlini2016robustness}, the algorithm presented will not begin a magnitude refinement before the target classification error is reached.  The threshold at which \cref{alg:meth:adv:all} switches between minimizing the correct class' post-softmax prediction $s_t$ and minimizing the attack magnitude is defined by $g(s, t)$.

\alg[label=alg:meth:adv:all]
    {Process used to generate adversarial examples.}
    {
      \KwIn{%
        $N$, the number of attack-optimizing steps; $f(\cdot)$, the NN; $x$, the network input; $t$, the true class of the input; $o(\cdot)$, an optimizing method such as SGD with momentum; $g(s, t)$, a goal function returning true if the network outputs from the attack are suitably different from the true class $t$; $\eta$, a balancing term between categorical loss and MSE loss.
      }
      \KwOut{$\delta_{best}$, the adversarial noise which satisfies the goal $g(\cdot)$ and has minimal vector length.}
      \Begin{
        $\delta \gets \vec{0}$ \\
        $M_{best} \gets \inf$ \\
        \For{$n \in [0, ..., N-1]$}{
          $\hat{x} \gets c(x + \delta)$ \tcp*[h]{$c(\cdot)$ clips elements of its argument to a valid input range, e.g. $[0, 1]$} \\
          $y \gets f(\hat{x})$ \\
          $s \gets softmax(y)$ \\
          \If{$g(s, t)$}{
            $\Delta \delta \gets 2\delta$ \tcp*[h]{$L_2$ loss for magnitude} \\
            \If{$||\delta||_2 < M_{best}$}{
                $M_{best} \gets ||\delta||_2$ \\
                $\delta_{best} \gets \delta$ \\
            }
          }
          \Else{
            $\Delta \delta \gets \partial s_t / \partial (x + \delta)$ \\
            $\Delta \delta \gets \eta \frac{\Delta \delta}{||\Delta \delta||_2}$ \tcp*[h]{Fixed gradient magnitude}
          }
          $\delta \gets o(\delta, \Delta \delta)$ \tcp*[h]{Apply optimizer step}
        }
      }
    }

We present two choices of $g(s, t)$ for the current work.  The first, $g_{adv}(s, t)$, was the well-known adversarial attack metric used by all prior work in this field \cite{Tsipras2018robustness,Cohen2019certified}, and denotes the boundary at which top-1 accuracy would decrease:

\begin{align}
    g_{adv}(s, t) &= \begin{cases}1 & \text{if } s_t - max_j(s_j; j \ne t) < 0, \\
            0 & \text{otherwise}.
            \end{cases}
\end{align}

This was the $g(s, t)$ used to produce \cref{fig:research:acc-vs-rmse}.  When ARA values are reported for a model, we evaluated random validation or testing images until we had \SI{1000}{} which were correctly classified.  We then made a list of the RMSEs below which each image would retain the correct classification, minimized as per \cref{alg:meth:adv:all}.  This list was extended with $0$s for each image evaluated which was initially incorrectly classified: if a model scored \SI{70}{\%} classification accuracy on unmodified images, we would have a final RMSE list of about \SI{1429}{} in length, \SI{1000}{} of which were non-zero.  This list was then evaluated for accuracy at different levels of RMSE, as seen in \cref{fig:research:acc-vs-rmse}, and the area above the naive baseline was taken to produce the attack ARA metric.

In the context of \cref{alg:meth:adv:all}, we used $N=450$, $o(\cdot)$ was a {\em Stochastic Gradient Descent} (SGD) optimizer with a learning rate of $0.01$ and momentum $0.9$, and $\eta = 0.55$.  Examples of our attack against a regular JSRT network can be seen in \cref{fig:related:adv}, and against a regular CIFAR-10 ResNet-44 network in \cref{fig:results:adv-compare}.

\subsubsection{Adversarial Attacks as Explanations}\label{sec:meth:adv:explain}
In the context of explanations, however, we found the $g_{adv}(s, t)$ metric to be lacking.  The decision boundary was not always sufficiently distant from the data point to reveal salient features.  Instead, we targeted an amount of perceptual difference between the explanation and the original input, optimizing the shape of the perturbation for that which would maximally impact the network's output in a desired manner.  Comparing these explanations with the original input then demonstrates precisely which features would lead to a desired output.  This was accomplished by following $\partial s_t / \partial(x + \delta)$ up to a boundary RMSE, at which point the RMSE would be minimized, a tick-tock method similar to \cref{alg:meth:adv:all}, but substituting a slightly different boundary criteria:

\fig[label=fig:meth:adv:explain:plus-vs-minus]
    {\graphic{plots/explain-plus-vs-minus.pdf}}
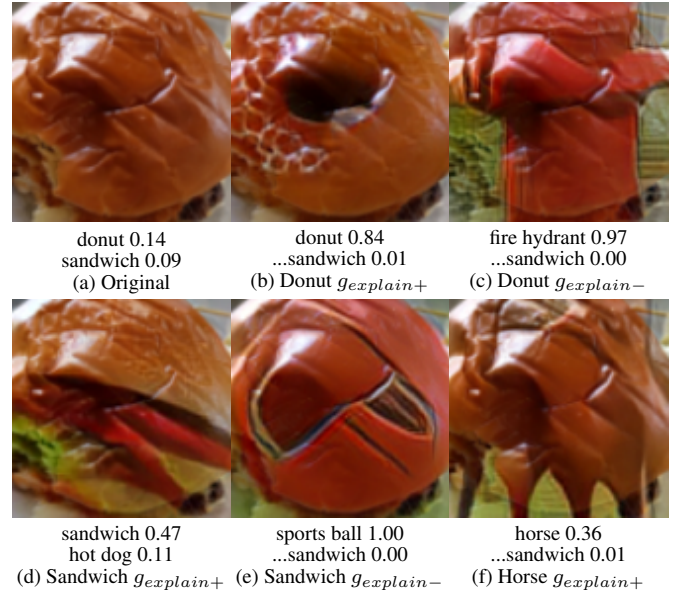
    {Different explanation techniques using $\rho=0.075$.  (a) The original image.  (b) A positive explanation for the donut class; we note alignment of the added ``hole'' with a wrinkle in the original sandwich bun.  (c) A negative explanation for the donus class resulted in the removal of the round shape of the sandwich.  (d) A positive explanation for the true class, sandwich, results in exposed contents (peppers or tomatoes), and the beginnings of lettuce.  (e) A negative explanation for the sandwich class reveals homogenization of the bun's texture, and further rounding out of the overall shape.  (f) Positive explanation for a completely unrelated class, horse: legs were clearly added, and the textured area in the upper-left of the image is appropriated as a saddle.}

\begin{align}
    g_{explain+}(\delta; \rho) &= \begin{cases}1 & \text{if } ||\delta|| > \rho, \\
            0 & \text{otherwise}.
            \end{cases}
\end{align}

We note that $g_{explain-}$ is also possible, by modifying \cref{alg:meth:adv:all} to maximize the selected class loss rather than minimizing it.  These techniques are demonstrated in \cref{fig:meth:adv:explain:plus-vs-minus}.  A more detailed analysis of interpreting the resulting AE images is provided in the results, \cref{sec:results:cifar-explain}.

Quantitatively, the attack ARA was not found to be indicative of the quality of these explanations.  For example, consider two closely related classes from CIFAR-10: automobile and truck.  These classes are often confused for one another, leading to a decrease in the magnitude of untargeted attacks for members of either class.  With respect to the network's ability to tell these two apart, $g_{adv}$ remains a good metric.  However, as a classifier learns to distinguish these related classes from the other unrelated classes, the $s_j$ corresponding to these related classes might rise in tandem.  The described phenomenon is illustrated in \cref{fig:meth:adv:explain:acc-vs-btr}.  This situation would indicate that the network possesses a greater capacity for deciding what is ``automobile'' or ``truck'' compared to the remaining classes, but the attack magnitude would not decrease as these two classes would still be easily confused.  Since the confusion between these two classes is built into the problem, $g_{adv}$ hits a ceiling beyond which an attack magnitude based on the $g_{adv}$ metric cannot be improved.  As such, we also considered {\em Better Than Random} (BTR) as a measure of the classifier's knowledge of class-specific features.  The BTR magnitudes were defined based on the distance between the classifier's prediction and a prediction at which the true label's valuation matches that of a random classifier.  As shown in \cref{fig:meth:adv:explain:acc-vs-btr}, the BTR quantity continues to increase even as related classes both become more confident predictions.  Thus, $g_{btr}$ (where $V$ is the number of classes in the prediction) is defined as:

\fig[label=fig:meth:adv:explain:acc-vs-btr]
    {\graphic{figs/acc-vs-btr.svg}}
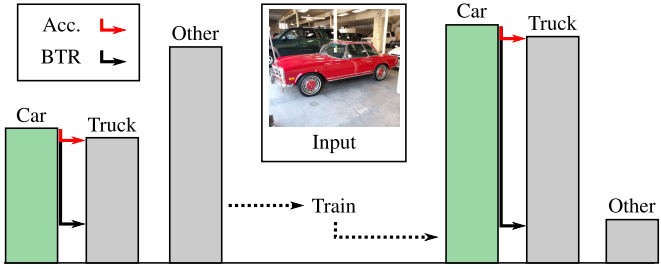
    {The previously mentioned attack ARA metric depends on the relative confidence between the correct class and the second-most-confident class.  For related classes such as ``car'' and ``truck,'' the distance between these two classes may not increase through training.  However, by measuring the distance from the correct class back to a fixed baseline, as is done with the BTR ARA metric, improvements in feature recognition may be measured regardless of the presence of related classes.}

\begin{align}
    g_{btr}(s, t) &= \begin{cases}1 & \text{if } s_t < \frac{1}{V}, \\
            0 & \text{otherwise}.
            \end{cases}
\end{align}

We note that the numerical stability of BTR is guaranteed, as resetting all pre-softmax outputs to $0$ achieves the required condition.  The BTR ARA gracefully degrades into that attack ARA on binary classification problems.  Note also that we deliberately chose a truly random classifier, and not a naive classifier, for unbalanced datasets (such as the Microsoft COCO dataset, \cref{sec:meth:arch:coco}).  When calculating BTR ARA metrics from a population of adversarial examples created using $g_{btr}$, a naive classifier was still used as the baseline for the area calculation.

\subsubsection{Example ARA Metrics}

For a traditional ImageNet ResNet-18, we measured an attack ARA of \valImagenetOurAucTraditional\ and a BTR ARA of \valImagenetOurBtrAucTraditional.  For a CIFAR-10 ResNet-44, we measured an attack ARA of \valCifarOurAucTraditional\ and a BTR ARA of \valCifarOurBtrAucTraditional.  An intuitive sense of attack ARAs may be gathered from \cref{app:attack-ara}, and BTR ARAs are compared in \cref{app:cifar,app:coco,app:imagenet}.

\subsection{Defense via Lipschitz Continuity}\label{sec:meth:lipschitz}
An integral part of many white-box attacks, including \cref{alg:meth:adv:all}, involves following the gradient of some loss.  The rate at which the output of the network might be changed is likewise dependent on that gradient.  To see how this might affect classification networks, consider the softmax operation, here denoted as $s(\cdot)$, applied to the output of an NN, $y$:

\begin{align}
  s(y) &= \frac{e^{y}}{\sum_{i=1}^V{e^{y_i}}}. \label{eq:softmax}
\end{align}

In the \SI{1000}{}-class ImageNet Large Scale Visual Recognition Competition 2012 (ILSVRC 2012) challenge, there are \SI{1000}{} classes \cite{ILSVRC15}.  Assuming 999 of those classes have an NN output of $y_i = 0$, then a value for the remaining class of $y_j = 10$ corresponds to a confidence in class $j$ of \SI{95.7}{\%}.  For a confidence of \SI{4.3}{\%}, that value need only fall to $y_j = 3.8$.  In reality, an adversarial attack also has the option of increasing the confidence of classes $i\ne j$ to reduce confidence of class $j$.  If $y_j = 3.8$, and another $y_k = 6.2, k\ne j$, then the confidence of class $j$ falls to \SI{2.9}{\%} and class $k$ skyrockets to \SI{32.1}{\%}.  In other words, instability on the output values will quickly overwhelm the softmax operation.  If we assume locally-linear behavior of the network, this instability may be modeled by looking at the expected change in the network's output given some gradient information.  Using $E_i\left[\cdot\right]$ to denote an expectation conditioned on $i$, $N$ as the number of input elements, $\Delta$ to signify an actual value change, and $\partial$ to signify a variable's partial:

\begin{align}
E_i\left[|\Delta y_i|\right] &\approx E_i\left[\left|\sum_{j=1}^N \Delta x_j \frac{\partial y_i}{\partial x_j}\right|\right], \\
    &\lessapprox \sum_{j=1}^N E_{i,j}\left[ \left| \Delta x_j \frac{\partial y_i}{\partial x_j}\right|\right], \\
    \text{These quantities are neither independent nor equivalent, } \span\nonumber\\
    \text{but we will simplify them as such:} \span\nonumber\\
E\left[|\Delta y_i|\right] &\lessapprox NE\left[\left|\Delta x_j\right|\right]E\left[\left|\frac{\partial y_i}{\partial x_j}\right|\right]. \label{eq:meth:attack}
\end{align}

\Cref{eq:meth:attack} provides a loose guideline for targeting different values of $|\partial y_i / \partial_x|$.  In fact, as a network becomes more non-linear, \cref{eq:meth:attack} becomes less accurate.

To see how effective the guideline given by \cref{eq:meth:attack} was in practice, we built a ResNet-18 and trained it on ILSVRC 2012 training data, detailed in \cref{sec:meth:arch:imagenet}.  Leveraging PyTorch's automated differential engine, we collected gradients for one of the NN's outputs, before the softmax, with respect to each of the \SI{150528}{} input elements ($224\cdot 224\cdot 3$).  The mean absolute value of the computed derivatives then resulted in an aggregate number which summarized the network's volatility in the original input space.  For our ResNet-18, this value worked out to \SI{0.051}{\per input}.  Interestingly, the mean of the maximum absolute derivative per image was a much larger \SI{3.9}{\per input}, indicating a large spread in these values.  Attacks were generated against this network with a \SI{50}{\%} confidence margin in favor of an adversarial class.  Again, based on a local-linearity assumption, the magnitudes of these attacks were measured as the mean absolute difference per pixel between the original and attacked images.  The harmonic mean of the mean absolute distances of all such attacks against this network was found to be \SI{0.0033}{\per input}; according to \cref{eq:meth:attack}, the sum of $\Delta y_i$ between the true and adversarially targeted classes should then be less than \SI{50.7}{}.  The actual measured sum of $\Delta y_i$ across the true and target classes averaged \SI{26.1}{}.


The change in network output was shown in \cref{eq:meth:attack} to be bounded proportionally to the gradient of the output with respect to each input element, as long as local network behavior was linear.  Since this assumption seemed to hold for real networks, we theorized that limiting this gradient would therefore provide some adversarial resistance in these linear regions of the network by forcing additional non-linearities to compensate for the limitation.  This is a form of Lipschitz continuity, as discussed in \cref{sec:related:lipschitz}.  From another point of view, limiting $|\partial y_i / \partial x_j|$ makes each training element a stable point for the network, enforcing a neighborhood of validity for each decision.  The classification loss then enforces necessary non-linearities between these stable regions.  As such, this work's primary contribution is to explore the relation between limiting $E[|\partial y_i / \partial x_j|]$ and adversarial attacks.  In the context of \cref{eq:madry}, this moves the focus from attempting to solve the outer minimization equation directly to instead limiting the effects of traveling in the allowed attack space $S$.  We note that, particularly with the {\em Rectified Linear Unit} (ReLU) activation function, even a gradient of $\vec{0}$ does not guarantee a neighborhood of validity; see \cref{sec:meth:adv-train} concerning that issue.

For networks with \SI{1000}{} outputs, minimizing $|\partial y_i / \partial x_j|$ directly for all $i$ is computationally prohibitive - each training image processed would require \SI{1000}{} additional gradient propagations.  Instead, we use a regularizing loss which is stochastically defined with a scaling parameter $\psi$ and a power factor $z$:

\begin{align}
  v_k &\sim [1, 2, ..., V], \nonumber\\
  L_{adv,z} &= \frac{\psi}{KN}\sum_{k=1}^K \sum_{j=1}^N \left|\frac{\partial y_{v_k}}{\partial x_j}\right|^z. \label{eq:meth:lipschitz-loss}
\end{align}

\Cref{eq:meth:lipschitz-loss} therefore draws $K$ random indices (without replacement) from the available output nodes and minimizes the derivative of each selected output with respect to all inputs.  Backpropagation makes this an efficient computation regardless of the number of input elements.  When $K = V$, $L_{adv,z}$ ceases to be stochastic.  $K$ and $N$ are both included in the denominator such that the expected force per image relative to the classification loss is maintained regardless of the number of inputs or outputs.  \Cref{sec:results:cifar:psi,sec:results:cifar:k} demonstrates the effects both of the relative strength of this loss, through changing $\psi$, and by varying its stochasticity, through changing $K$.

In addition to investigating the absolute value form of \cref{eq:meth:lipschitz-loss}, using $|\partial y_{n_k} / \partial x_j|^z$, we investigated instead minimizing $L_{adv,z,q} = \left(\sum_{l=1}^N |\partial y_{n_k} / \partial x_l|\right)^q |\partial y_{n_k} / \partial x_j|^z$ for some values of $z$ and $q$.  We included these to illustrate that the proposed regularization technique is in fact a rich family of techniques based on approximations of which quantities are relevant for adversarial defense; a limited investigation of these metaparameters is found in \cref{sec:results:cifar:adapt-ladv}.  From this point forward we will use $L_{adv}$ to refer to any of these, with default values of $z=2$ and $q=0$ unless otherwise specified.

We also considered the effects of creating a ``dead zone'' where gradients would not be punished, like a true Lipschitz constraint.  For these experiments, instead of minimizing based on $|\partial y_{n_k} / \partial x_j|$ directly, $L_{adv}$ would be minimized based on $max(|\partial y_{n_k} / \partial x_j| - \delta, 0)$.  Results are found in \cref{sec:results:cifar:deadzone}.

It is also possible for $n_k$ to be drawn from a non-uniform distribution.  To test the merits of this, we considered distributions which yielded the correct label $\zeta\%$ of the time and were pulled from a random distribution (including the correct label) the rest of the time.  Results with this technique are discussed in \cref{sec:results:cifar:zeta,sec:results:cifar:adapt-zeta}.

Another variant of non-uniform distribution involved substituting the minimization of the true class' gradient $\zeta\%$ of the time for minimizing the gradient $(\partial y_t - max_{i \ne t} \partial y_i) / \partial x_j$, the difference between the true class and the maximum non-true class prediction.  This regularization, which we label $L_{adv,tandem}$ because it aligns the gradients of two different classes in tandem, was chosen based on the ``automobile'' vs ``truck'' discussion from \cref{sec:meth:adv:explain}.  While regularizing only one class at a time guarantees that the gradient for that class will approach zero, this provides an opportunity for a related class to dominate.  Since the softmax operation assigns probabilities based on the difference between elements of its input, it was determined that it might be more effective to regularize the difference between those inputs (the NN's output).  Results for this technique are presented in \cref{sec:results:cifar:adapt-zeta}.

\subsection{Gradient Minimization as Weight Regularization}\label{sec:meth:weightreg}
Exploring analogs to minimizing $|\partial y_i / \partial x_j|$ further, consider a single layer of an NN:

\begin{align}
  \vec{y} &= f(\mathbf{A}\vec{x} + \vec{b}), \nonumber\\
  \frac{\partial{\vec{y}}}{\partial \vec{x}} &= \frac{\partial f(\cdot)}{\partial \cdot} \mathbf{A}.
\end{align}

Assuming all paths are active and we're using a ReLU network, then $\partial f(\cdot) / \partial \cdot = 1$.  Since we would then use the element-wise absolute value or square of each element of $\mathbf{A}$ to devise our adversarial loss function $L_{adv}$, this is identical to $L_1$ or $L_2$ regularization for $q=0, z\in\{1,2\}$.  While in a multi-layer setup, the proposed $L_{adv}$ diverges from standard weight regularization, we considered it worthwhile to run experiments with weight regularization disabled on the convolutional weights in the network (keeping it enabled on biases within the network).  These are explored in \cref{sec:results:cifar:weightreg}.

\subsection{Half-Huber Rectified Linear Unit (HHReLU)}\label{sec:meth:relu2b}
A classical ReLU is continuous in value, but its derivative is discontinuous.  Our proposal required optimizing the derivative of the activation functions used by the network, and as such we desired the first derivative to be continuous, allowing that the second derivative might be discontinuous.  Related to the Huber function, we devised a new activation function, the {\em Half-Huber ReLU} (HHReLU), defined as:

\begin{align}
  f(x) &= \begin{cases}
    0, & x < 0, \\
    dx^2, & x < \frac{1}{2d} \\
    x - \frac{1}{4d}, & \text{otherwise}.
  \end{cases} \label{eq:meth:relu2b}
\end{align}

While the parameter $d$ describes the acceleration of the $x^2$ region, for timely results, we did not explore this parameter outside of $d=1$.  Nonetheless, we note that other values of $d$ or other activation functions with a continuous first derivative might be explored further in the future.  The impact of using this activation function instead of a traditional ReLU is explored in \cref{sec:results:cifar:relu2b}.

\subsection{Output Zeroing}\label{sec:meth:outzero}

The softmax function, \cref{eq:softmax}, is translation-invariant with respect to its inputs.  We found that, in practice, allowing networks to rely on the invariance of the softmax function resulted in the flattening effects of $L_{adv}$ persisting classification errors within the network.  For instance, assume a two-class NN, which is producing output $y_0 = 1, y_1 = 2$ for some input.  The adversarial resistance loss from \cref{eq:meth:lipschitz-loss} induces a certain amount of inertia about $y_0 = 1, y_1 = 2$, making it harder for the network to switch the ordering of these outputs.  Adding an additional $L_2$ regularization term for the pre-softmax output of the network biases all of these terms toward $0$, easing the classification task:

\begin{align}
  L_{out} &= k_{out}\sum_i y_i^2.
\end{align}

If $k_{out}$ is too large, then the network will never gain any confidence in its answers.  Too small a $k_{out}$ and the benefits will disappear.  Therefore, similar to the guidelines on $|\partial y_i / \partial x_i|$ established by \cref{eq:meth:attack}, we provide some guideline calculations regarding the balancing of these two forces while considering the maximum learnable confidence when training with a cross-entropy loss $L(\cdot)$ (using $V$ as the number of possible classes):

\begin{align}
  \text{Assume $y_z = 0$, where $z \ne t$}, \span\nonumber\\
  L(\vec{y}) &= -log(s_t), s_i = \frac{e^{y_i}}{e^{y_i} + V - 1}, \nonumber\\
  dL/dy_t &= 
      s_t - 1.
\end{align}

\noindent When the network's predictions are an accurate distribution, and examples are uniformly distributed, $y_i$ only has a $1 / V$ chance of being a large value and needing to contest the classification loss.  The rest of the time, it would only have a $1 - s_i$ chance of being large for a confusing example.  Balancing the force of the cross-entropy loss, $L(\cdot)$, with $L_{out}$ yields:

\begin{align}
  \frac{L}{V} &= \frac{L_{out}}{V} + \frac{(V - 1)(1 - s_t)L_{out}}{V}, \nonumber\\
  \frac{1 - s_t}{V} &= \frac{k_{out}2y_t}{V} + \frac{(V - 1)(1 - s_t)k_{out}2y_t}{V}, \nonumber\\
  k_{out} &= \frac{1 - s_t}{2y_t(1 + (V - 1)(1 - s_t))}. \label{eq:meth:kout}
\end{align}

\Cref{eq:meth:kout}, like \cref{eq:meth:attack}, is not claimed to be an exact equation.  However, it gives a guideline for reasonable parameter values.  We tried $s_t = 0.8$ for all experiments, yielding $k_{out} = 0.01$ for our CIFAR-10 experiments, $k_{out} = 6e-5$ for our ImageNet experiments, and $k_{out} = 1e-3$ for our COCO experiments.  Results of varying this parameter on the CIFAR-10 dataset may be found in \cref{sec:results:cifar:outzero}.

\subsection{Adaptive $\psi$}\label{sec:meth:adaptive}

To ease comparisons between the meta-parameters necessary for our proposed technique to work, we investigated setting $\psi$ from \cref{eq:meth:lipschitz-loss} automatically based on a targeted training loss.  We performed experiments using an integrating controller:

\begin{align}
  \psi &= k_{\psi,0} e^{k_{\psi}\sum_b c(-\ln(L_{train,b} / L_{target}))}, \label{eq:meth:adaptive:psi}\\
  c(d) &= clip(d, -\epsilon_{worse}, \epsilon_{better}), \nonumber
\end{align}

\noindent where $b$ is the batch index, $L_{target}$ is the targeted training loss, and $L_{train,b}$ is the training loss for batch $b$.  While this approach still has one significant metaparameter, $L_{target}$, the meaning of its value is consistent regardless of other parameters.  The other metaparameters for an adaptive $\psi$ were important only for guaranteeing that $\psi$ changed slowly, over a large number of batches.  In all experiments, the regularization proposed in this work was capable of matching $L_{target}$.  Note that the strength of $\psi$ is based on the exponential of the integral as we found this to work significantly better across different scales of $L_{target}$, and throughout network training.  The inner summation of \cref{eq:meth:adaptive:psi} was always prevented from falling below zero, making $k_{\psi,0}$ the minimum strength.

Values used were typically $k_{\psi,0}=220$, $k_{\psi} = 0.02$, $\epsilon_{better} = 1$, $\epsilon_{worse} = 0.01$, though early experiments used $k_{\psi,0}=0.01, \epsilon_{worse}=1$ (experiments before those discussed in \cref{sec:results:cifar:adapt-zeta}).

\subsection{Adversarial and Noisy Training}\label{sec:meth:adv-train}
\Cref{sec:meth:lipschitz} mentioned that, even with gradients of $\vec{0}$ at all points in the training data, an activation function like ReLU guarantees no neighborhood for which the gradient will remain $\vec{0}$.  That is, a loss ``cliff'' might be arbitrarily close to any training point.  This is partly due to the overparameterization of NNs, illustrated in \cref{fig:meth:adv-train:example}.  As such, we also investigated combining our method with either random Gaussian noise or adversarial training.

\fig[label=fig:meth:adv-train:example]
    {\graphic{figs/noisy-training.svg}}
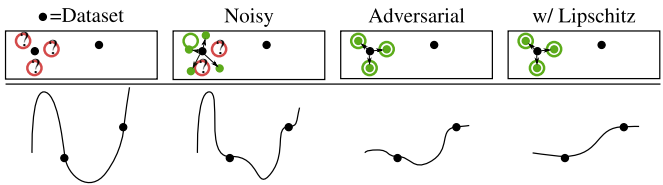
    {Training an NN on samples from a dataset (left) specifies desired behavior at the data points, but does not describe behavior in between those data points, allowing the decision surface to take an arbitrary shape.  Adding some noise (second from left) can somewhat improve the behavior between dataset samples, but is statistically unlikely to improve the worst behaviors.  Adversarial training (second from right) deliberately attempts to improve the worst performing points, leading to a smoother decision surface.  Adding a stochastic Lipschitz loss, as \cref{eq:meth:lipschitz-loss}, further smooths behavior between data points.}

Adversarial training was implemented two ways.  In the first way, denoted as $L_2$, an $L_2$ distance $\epsilon$ was chosen and the adversary attempted to find the highest loss value within that $\epsilon$-ball.  The gradient of the classification cross-entropy loss was followed for 7 steps, each time being normalized to $\epsilon / 7$ magnitude.  This was very similar to the original adversarial training approach proposed by Madry \etal\ \cite{Madry2017towardsResistantAdv}.  In the second way, minimal adversarial training denoted as $L_{2,min}$, an $L_2$ distance $\epsilon$ was also chosen, but before taking each step, the network's classification was evaluated.  If the network correctly classified the example, then the gradient of classification loss was normalized to length $\psi_{7,n}\epsilon$ and followed, where $\psi_{k, n} = 2(k - n) / [k(k + 1)]$ and $n \in [0, 1, ..., k-1]$ is the index of the step being taken.  In this formulation, the step sizes at subsequent steps yield progressively finer movements.  If the network incorrectly classified the example, then the gradient was replaced with the negation of the current perturbation, normalized to size $\psi_{7,n}\epsilon$, and followed.  That is, the $L_{2,min}$ method of adversarial training sought to train on adversarial examples near the boundary at which the network would misclassify those examples.

Training configurations where batches were composed of half adversarial examples and half original examples from the dataset were also considered.  In Tsipras \etal\ \cite{Tsipras2018robustness}, this technique was called ``Half-Half'' training, and we keep that nomenclature.

Neither of these guarantee that a loss cliff would be corrected, but as seen in \cref{sec:results:cifar:adv-train}, they both somewhat alleviate the underlying problem.  We also note that noise from batch normalization and data augmentation should help with this problem.

\subsection{Active Learning}\label{sec:meth:hitl}
We explored using the adversarial examples generated from our networks to bootstrap even better adversarial resistance in an active learning pipeline with a {\em User Interface} (UI) as shown in \cref{fig:results:hitl-ui}.  The UI took a trained network and used it to generate high-confidence adversarial examples.  These examples were also generated from \cref{alg:meth:adv:all}, but instead with a high-confidence condition of $g(s, t) = max_j(s_j; j \ne t) - max_q(s_q; q \ne j) > 0.5$.  That is, the adversarially generated, incorrect class had to be \SI{50}{\%} more confident than the next-highest class.  Users were then presented with three options: unchanged, unsure, and no longer the original class.  When any button was pressed, the adversarial image was saved along with the original label and the annotation.  We then tested re-training networks from scratch using the adversarial images annotated as ``unchanged'' as part of the training data.  While this method of feedback was somewhat limited, we offer it as a proof-of-concept that our method of producing adversarial explanations could be used not only to inform the user about the reasoning behind an algorithmic decision, but also to feed annotations back into improving the classifier.

\fig[label=fig:results:hitl-ui]
  {\graphic[width=\linewidth]{figs/hitl-ui.svg}}
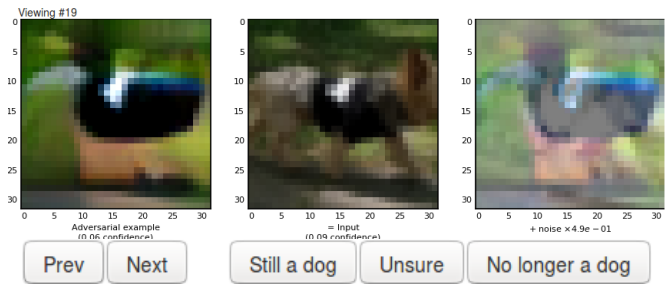
  {Prototype active learning interface.  Whereas adversarial training simply generates adversarial examples and trains them to be recognized as the original class, our adversarial explanations can be used for the generation of sufficient quality samples to merit human intervention as to whether or not the class of the image has changed.}

\subsection{Architectures and Datasets}\label{sec:meth:arch}

This section contains details on the architectures used for the various datasets discussed in this document.

Note that all networks were evaluated based on their final state during training - no validation set was used to cherry-pick peak performance during training.  Overfitting was not found to be a problem in the traditional sense, a result consistent with the original findings on residual networks \cite{He2016resnet}.

\subsubsection{CIFAR-10}\label{sec:meth:arch:cifar10}
CIFAR-10 is a commonly-used dataset with \SI{50000}{} training images and \SI{10000}{} test images, consisting of $32\times 32$ RGB images belonging to one of \SI{10}{} classes \cite{Krizhevsky2009cifar}.  Our CIFAR-10 experiments were based on a ResNet-44 \cite{He2016resnet}, modified to be in pre-activation form \cite{He2016preact} with each residual block's output convolution weights initialized to zero as per \cite{He2018bagOfTricks}.  Training used mini-batches of size 256 spread across 2 GPUs, for 128 images per GPU.  We used standard data augmentation techniques for this task, reflecting the bordering 4 pixels and taking a random $32\times 32$ crop during training.  Training images were horizontally flipped \SI{50}{\%} of the time.  {\em Stochastic Gradient Descent} (SGD) was used to optimize weights with a momentum of $0.9$ and $L_2$ weight regularization with a strength of $1e-4$, starting at a learning rate of $0.02$ which was linearly increased to $0.2$ over the first 10 epochs.  The learning rate was then stepped down to $0.02$ and $0.002$ at $170$ and $195$ epochs, respectively.  Training was halted at $200$ epochs.  This entire setup was implemented in PyTorch \cite{Paszke2017pytorch}, and resulted in a final top-1 validation accuracy of \SI{92.2}{\%} on a network without other changes.

\subsubsection{ILSVRC 2012}\label{sec:meth:arch:imagenet}
While CIFAR-10 is small enough to iterate on quickly, success on CIFAR-10 does not guarantee the generality of a technique.  Therefore, we also investigated training on the ILSVRC 2012 dataset, consisting of \SI{1281167}{} training and \SI{50000}{} validation RGB images of varying size but significantly higher resolution than CIFAR-10, with objects belonging to one of \SI{1000}{} classes \cite{ILSVRC15}.  We trained a ResNet-18 \cite{He2016resnet} modified to be in pre-activation form \cite{He2016preact} with each residual block's output convolution weights initialized to zero as per \cite{He2018bagOfTricks}.  To ease gradient descent with respect to the input as discussed in \cref{sec:meth:lipschitz}, we also replaced the initial max pooling operation with an average pooling operation.  $L_2$ regularization was applied to weights and biases with a strength of 1e-4 .  Training used mini-batches of size 192 spread across 3 GPUs, for 64 images per GPU.  We used standard data augmentation techniques for this task, resizing the smallest edge of each image in $[256, 480]$ and taking a random $224\times 224$ crop.  Each crop was then given a \SI{50}{\%} chance of being horizontally flipped.  We skipped the standard color augmentations.  Rather than using the state-of-the-art method of computing validation accuracy, which would have involved a 10-crop on the validation phase \cite{He2016resnet}, we instead resized images such that the smallest edge was $256$ pixels across and then took a crop of the central $224\times 224$ pixels for validation.  SGD was used to optimize weights with a momentum of $0.9$ and $L_2$ weight regularization with a strength of $1e-4$, starting at a learning rate of $0.03$ and linearly increased to $0.3$ over the first 10 epochs.  The learning rate was then stepped down to $0.03$ and $0.003$ at $50$ and $60$ epochs, respectively.  Training was halted at $65$ epochs.  This entire setup was implemented in PyTorch \cite{Paszke2017pytorch}, and resulted in a final top-1 validation accuracy of \SI{65.6}{\%} on a network without other changes.

\subsubsection{Microsoft COCO}\label{sec:meth:arch:coco}
The {\em Common Objects in COntext} (COCO) dataset \cite{Lin2014cocoDataset}\ was used as an additional proof-of-concept.  The dataset consists of images containing scenes of multiple annotated objects from 80 different classes.  To stick to classification problems for demonstrating our methods, we created a sub-dataset from COCO which consisted of taking the bounding box of each annotated object as a separate input example.  During training, each object's sub-image was resized such that the smallest edge was between $96$ and $120$ pixels long, selected a random $96\times 96$ crop, and randomly performed a horizontal flip.  During validation, each object was resized such that the smallest edge was $108$ pixels long, and then the central crop of $96\times 96$ pixels was selected.  This scheme often led to images that overlap with the ``person'' classification, but was sufficient as a proof-of-concept.

The base network used for COCO annotations was the same ResNet-44 network from \cref{sec:meth:arch:cifar10} as used CIFAR-10, but with filters of size $32$, $64$, and $128$ (twice the standard width).  Rather than the standard convolution for transforming input data for the first residual block, we used a convolutional layer with a kernel size of 4 and a stride of 3, which reduced the image from $96\times 96$ to $32\times 32$.  SGD was used with a momentum of $0.9$ and $L_2$ weight regularization with a strength of $1e-4$, starting at a learning rate of $0.02$ which was linearly increased to $0.2$ over the first 10 epochs.  The learning rate was then stepped down to $0.02$ and $0.002$ at $55$ and $70$ epochs, respectively.  Training was halted at $80$ epochs.  This entire setup was implemented in PyTorch \cite{Paszke2017pytorch}, and resulted in a final top-1 validation accuracy of \SI{77.4}{\%} on a network without other changes.  We note that this accuracy was calculated on imbalanced data.  For example, the greatest imbalance in our validation dataset was for the class ``person,'' which accounted for \SI{31.4}{\%} of all objects in the dataset.  In the context of the other datasets used in this work, top-1 accuracy is an easily-understood metric.  We have therefore continued to use that metric on COCO.  The calculation of the attack ARA (\cref{sec:meth:adv:accuracy}) and BTR ARA (\cref{sec:meth:adv:explain}) specify uses of naive or random classifiers as appropriate to deal with the imbalance in the COCO dataset.

\subsubsection{JSRT}\label{sec:meth:arch:jsrt}
The JSRT was described in \cref{sec:related:expl}.  Our JSRT results were produced with networks similar to the ResNet-44 networks for CIFAR-10 from \cref{sec:meth:arch:cifar10}, using filters of size $64$, $96$, and $128$, and with an initial convolution of kernel size $9$ followed by an average pooling layer of size $8$.  Additionally, each input image was normalized such that it had zero-mean and unit variance; this was done due to wild variations in the different scans and scanned regions.  Regions of $256\times 256$ pixels were selected either A) with the central point being part of the nodule annotation for images containing nodules, or B) entirely randomly for images not containing nodules.  Malignant and benign classifications were considered the same, under a new ``nodule'' category (making the problem binary).  Training images were heavily augmented with shear angles from $[-30, 30]$ degrees, rotated from $[-45, 45]$ degrees, and scaled on a factor of $[0.61, 1.65]$.  Additionally, random square regions of the final training image between $[0, 64]$ pixels on each side were set to either black or white, to augment against the earlier per-image normalization.

\subsection{Code Availability}
A reference implementation of the techniques presented throughout this section applied to the CIFAR-10 dataset may be found at \url{https://github.com/wwoods/adversarial-explanations-cifar}.

\section{Results}\label{sec:results}

The majority of our experiments were conducted on CIFAR-10 due to it being a smaller dataset which was faster for iterating parameters and ideas.  The utility of explanations produced via AE is explored on CIFAR-10 in \cref{sec:results:cifar-explain}.  CIFAR-10 experiments detailing ablations of the methods from \cref{sec:meth} are explored in detail in \cref{sec:results:cifar}.  Experiments on ILSVRC 2012 were also conducted, and are covered in \cref{sec:results:imagenet}.  Experiments on the COCO dataset are covered in \cref{sec:results:coco}.


\subsection{CIFAR-10 Adversarial Explanations}\label{sec:results:cifar-explain}

\fig*[label=fig:results:cifar-explain:main]
  {\graphic{figs/explain-vs-gradcam-cifar-annotated.pdf}}
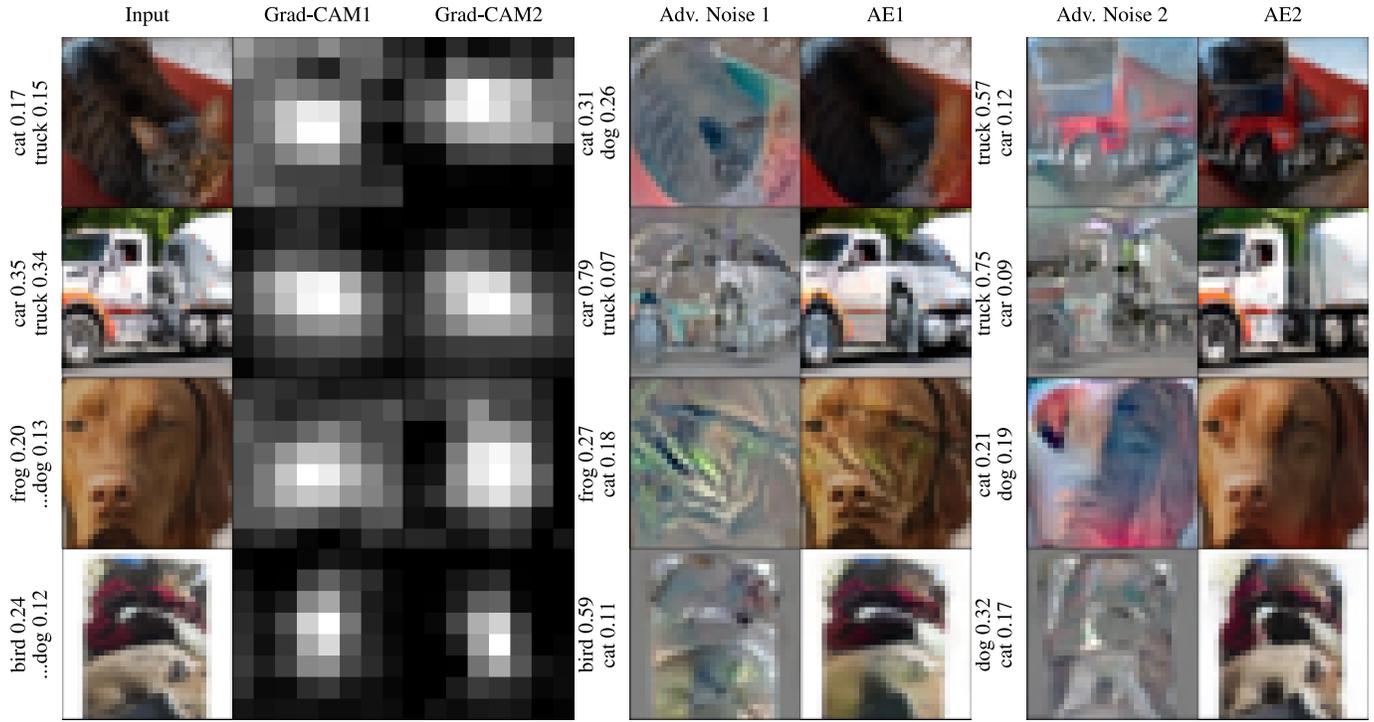
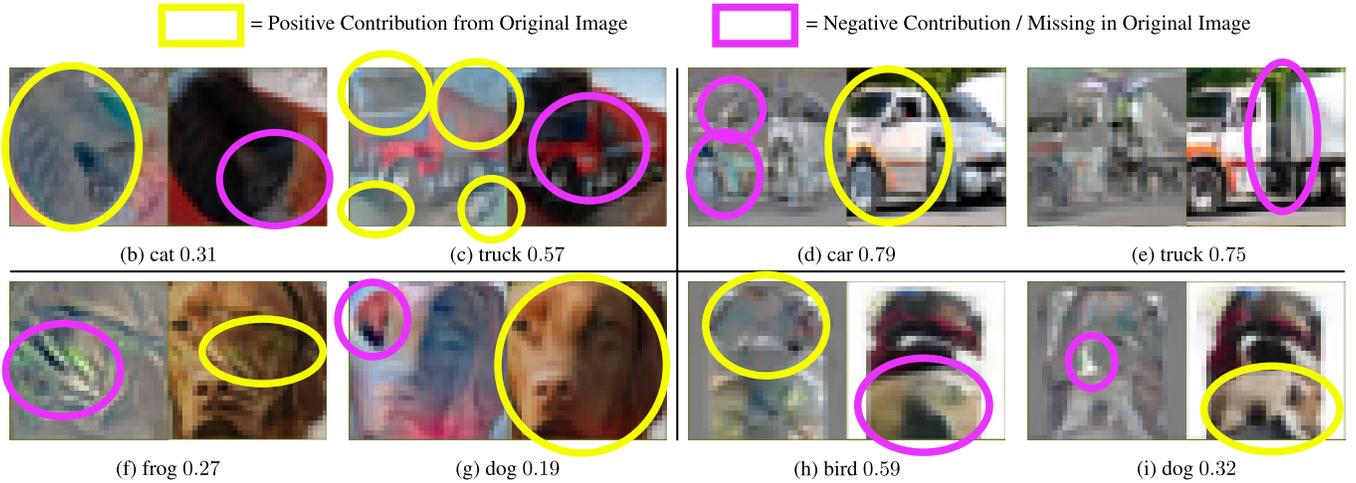
  {Exploring the explanatory power of AEs.  The top four rows, subfigure (a), are in the same format as \cref{fig:research:adv-motivation}.  Below that, (b) through (i) are annotated versions of the AEs for subfigure (a), indicating regions which contributed to or detracted from each predicted class.  See \cref{sec:results:cifar-explain} for full commentary.}

A comparison of the Grad-CAM method of explaining an NN and our AEs is shown in \cref{fig:results:cifar-explain:main}.  This figure was produced using our CIFAR-10 network with the highest attack ARA.  The left half of (a) demonstrates four different input images, and the corresponding NN predictions for the most confident class and either the second-most-confident class, or the true class, if it was not the most confident prediction.  Next to the input image are ``Grad-CAM1'' and ``Grad-CAM2,'' containing the Grad-CAM explanations for the two displayed class predictions.  We note that even for very disparate classes, such as ``cat'' and ``truck'' in the first row, the Grad-CAM explanations are mostly the same, and do little to indicate the textures or shapes which influenced the decision.  Following the Grad-CAM explanations, in the right half of (a), are two AEs, representing $g_{explain+}$ for each of the two class predictions displayed by the original input image.  Each AE shows the new top-two network predictions, an image of the differences between the original input and the adversarial image, and the adversarial image itself.  Below (a) are subfigures (b) through (i), which detail each of the AEs.

Subfigures (b) and (c) demonstrate relevant conclusions which may be drawn from the AEs in row 1 of (a).  The network correctly classified this image as ``cat,'' but from the difference image in (b), it can be seen that the ``cat'' class confidence would have been even higher with a blacker body and without the cat's face.  The body was annotated as a positive contribution because, while the adversarial image changed the body, it kept the overall structure of that region, and increased its contrast.  On the other hand, the cat's face is almost entirely removed from the adversarial image, indicating it contributed against the ``cat'' classification.  This indicates that the NN did not possess the logic needed to recognize a face in that configuration as belonging to a cat, perhaps because the cat's face is too small of a feature in the image.  In (c), the explanation for the ``truck'' prediction illustrates that the framing of the central cat mimics the framing of many ``truck'' photos in the training data.  That is, the shapes of the corners of the image were well preserved, with the high-contrast upper-right corner being similar to the division between a trailer and the sky.  The truck which was added as part of the explanation was missing in the original image, and was thus annotated as a counter-indicator.  Note though that the RMSE between the original image and either of (b) or (c) is the same - while the truck is a significant addition in terms of detail, the cat's body was preferentially much darker for a more confident ``cat'' classification.
Note also that the final class confidences for these AEs are $0.31$ and $0.57$, indicating that in $L_2$ space, the input image is much closer to a large ``truck'' manifold than a ``cat'' manifold.  With AEs, we gain information about the network's function not only through the input features which would be need to be modified, but also through the resulting class confidences.

Subfigures (d) and (e) annotate the AEs from row 2 of (a).  Interestingly in (d), the adversarial explanation for ``car'' relaxes the slope of the pillar against the windshield, and removes much of the coloring around the wheel well.  Neither of those features would often be found in cars, though they were present in the original input.  With these modifications in place, the shape of the vehicle's front matches that of a car, and it becomes unclear whether or not the trailer is in the foreground.  The ``truck'' AE, (e), indicates that the main reason this input was not identified as a ``truck'' was the missing gap between the tractor and the trailer.  With that feature in place, confidence in a ``truck'' class skyrocketed.

Subfigures (f) and (g) are the AEs corresponding to row 3 of (a).  The reasoning behind the network's final guess of ``frog'' was hard to see at first, but two major factors clearly contributed.  First, in (f), the frog-skin shading already existed on the right side of the face.  While the AE exaggerated this shading, it was clearly already present.  Second, in (g), more or less the entire image was turned more red and higher contrast to inspire a ``dog'' prediction.  Looking at the final confidences, with a maximum of $0.27$ even with significant perturbations, this image was likely somewhat distant from the original training data's manifold, and possessed just enough of the frog-skin shading on the face to convince the network of the ``frog'' class being most applicable.

Subfigures (h) and (i) are the AEs corresponding to row 4 of (a).  Here, in (h), the top half of the image was similar to a bird face when rounded out a bit.  The actual dog pixels in the bottom half of the image were significantly smoothed in this AE, indicating that they were counter-indicators of the ``bird'' class.  In (i), one key feature prevented a dog classification.  If the white piece of clothing in the original image's middle-left swept further down, then the center of the image would have looked more dog-like, with the resulting black bubble forming a nose.  It is also clear that a bit more contrast within the dog's pixels would also have helped.

Altogether, AEs show significantly more information about the NN's operation than prior state-of-the-art techniques like Grad-CAM.  Full-color, textured explanations in the form of AEs lend not only the ability to see which features are missing for a given classification, but also the effect that adding those features would have on the predicted class confidences.  Unlike previous approaches, AEs also take the NN's non-linearities into account.

\subsection{CIFAR-10 Experiments}\label{sec:results:cifar}

\definecolor{exp-original}{HTML}{440154}
\definecolor{exp-madry}{HTML}{31688E}
\definecolor{exp-madrymod}{HTML}{35B779}
\definecolor{exp-best}{HTML}{FDE725}
\def\colordot{\raisebox{-1.25pt}{\Large$\bullet$}}
\def\tcolordot{\raisebox{-1.1pt}{\Large$\bullet$}}

All CIFAR-10 experiments run with a ResNet-44 have been plotted in \cref{fig:results:adv-acc}.  At each level of accuracy (x axis), there may be several dots for ARA (y axis), indicating separate experiments with different levels of adversarial resistance.  The variance of individual experiments is indicated in \cref{sec:results:cifar:k}; generally, the standard deviation for any of final classification accuracies was $\pm$\SI{0.2}{\%}, and the standard deviation for ARA calculations was $\pm$\SI{0.0001}{}.  The most immediate quality to be seen comparing best-in-class RMSEs across different accuracies is that accuracy may be sacrificed for additional adversarial resistance.

Large, colored dots indicate selected experiments.  \textcolor{exp-original}{\tcolordot} N1 indicates a traditional ResNet-44, not modified for increased resistance.  \textcolor{exp-madry}{\tcolordot} N2 indicates a traditional ResNet-44 trained with adversarial training alone.  \textcolor{exp-madrymod}{\tcolordot} N3 indicates a ResNet-44 with the modifications from \cref{sec:meth}, and was trained with both adversarial training and $L_{target}=1.5$.  \textcolor{exp-best}{\tcolordot} N4 indicates a similar ResNet-44 to N3, but without the adversarial training.  A comparison of adversarial attacks against the networks indicated by colored dots may be found in \cref{fig:results:adv-compare}.  The adversarial attacks against even the most robust of the networks were still very small perturbations, but did result in the visible accentuation of certain features, particularly the car door.  An ideal network would produce genuine ambiguity at the adversarial example boundary.  However, the right side of this figure demonstrates the proposed explanation techniques applied to the different networks.  Each of the $g_{explain}$ columns were produced with an RMSE of $0.15$; N1's explanations look like static, indicative of that network's spurious decision-making boundaries.  While the example for the adversarially trained N2 is beginning to exhibit salient features, there is little difference between the ``car'' and ``cat'' columns.  In contrast, N3 and N4 both demonstrate clear features, illustrating the utility of our stochastic Lipschitz regularization and other techniques for producing networks capable of generating coherent explanations.

See \cref{app:cifar} for more examples of adversarial explanations with our CIFAR-10 networks.  We emphasize that adversarial attacks against networks using our regularization term demonstrated increasingly salient features from the targeted class as the BTR ARA metric increased.  These salient features were not forced from any term which necessitated a reconstruction of the input, as one would see with a GAN or VAE, indicating that the proposed technique alone was sufficient for producing classifiers which rely on salient features.

We note again that classifiers with a high clean accuracy and low attack ARA are more fragile classifiers.  The generalization difference between the \textcolor{exp-madry}{\tcolordot} N2 and \textcolor{exp-madrymod}{\tcolordot} N3 networks may look significant, with accuracy falling from \SI{87.4}{\%} to \SI{68.4}{\%}, but the ARA rose from \SI{0.0107}{} to \SI{0.0197}{}.  The \textcolor{exp-madry}{\tcolordot} N2 network loses all predictive power against an adversary with a maximum attack RMSE of $0.03$, while the \textcolor{exp-madrymod}{\tcolordot} N3 network retains its predictive power out to an RMSE of $0.07$, and the \textcolor{exp-madrymod}{\tcolordot} N3 network becomes the more accurate of the two at an RMSE of $0.01$.  An RMSE of $0.01$ on CIFAR-10 is a virtually imperceptible change; see \cref{app:cifar-4} for an example of different perturbation magnitudes.  \Cref{sec:results:cifar-explain} was produced using the \textcolor{exp-madrymod}{\tcolordot} N3 network.

\fig[label=fig:results:adv-acc]
  {\graphic{plots/all-experiments.pdf}}
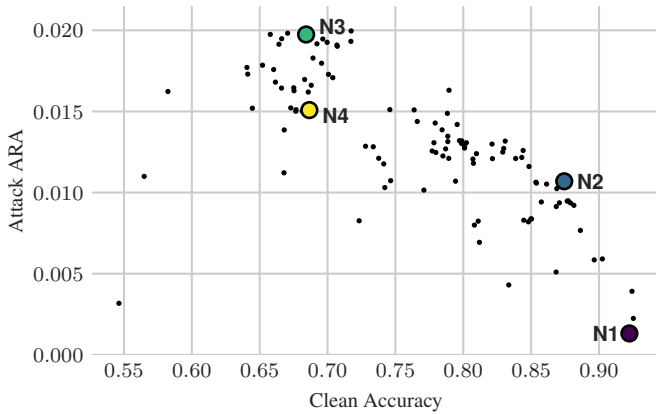
  {Plot of all CIFAR-10 experiments run with a ResNet-44; accuracy on clean data versus attack ARA (as per \cref{sec:meth:adv}).  \textcolor{exp-original}{\colordot} N1 indicates an unmodified ResNet-44, \textcolor{exp-madry}{\colordot} N2 indicates a network trained with only adversarial training, \textcolor{exp-madrymod}{\colordot} N3 indicates a network which combines adversarial training and our regularization, and \textcolor{exp-best}{\colordot} N4 indicates an experiment using only our regularization.  Dotted experiments are also used in \cref{fig:results:adv-compare}.}

\fig[label=fig:results:adv-compare]
  {%
    \graphic{plots/ex_cifar10.pdf}
  }
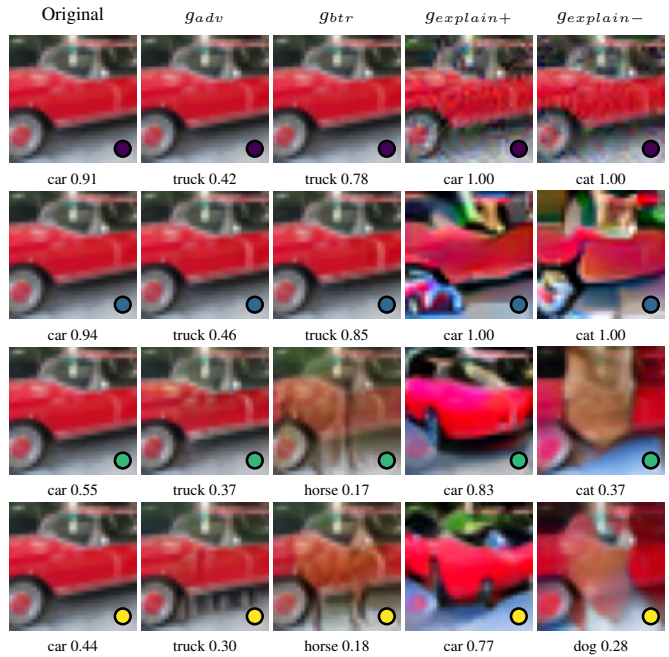
  {Input (left) and adversarial examples of different CIFAR-10 classes, generated for NNs with different levels of adversarial resistance using the different $g(\cdot)$ functions from \cref{sec:meth:adv}, as indicated by the heading at the top of each column.  The relative accuracy and attack RMSEs may be compared in \cref{fig:results:adv-acc}; from top to bottom, these correspond to the experiments denoted by \textcolor{exp-original}{\colordot} N1, \textcolor{exp-madry}{\colordot} N2, \textcolor{exp-madrymod}{\colordot} N3, and \textcolor{exp-best}{\colordot} N4 dots.  $g_{explain-}$ was generated by emphasizing the ``cat'' category.}

The most important CIFAR-10 experiments are detailed in \cref{tab:results:adv-mods-cifar,tab:results:adv-adapt-cifar}.  \Cref{tab:results:adv-mods-cifar} addresses ablations of the techniques mentioned in \cref{sec:meth:lipschitz,sec:meth:weightreg,sec:meth:relu2b,sec:meth:outzero}, showing that these techniques all work together to provide a reasonable level of adversarial resistance.  \Cref{tab:results:adv-adapt-cifar} illustrates the merits of \cref{sec:meth:adaptive,sec:meth:adv-train,sec:meth:hitl}.  The following sections share the same titles as the table entries for easy cross-referencing.  Where applicable, rows in the tables have the same colored dots as \cref{fig:results:adv-acc}, indicating the exact experiments conducted for those results.

\newcolumntype\wtableColumns{%
    >{\raggedright\hangindent=0.5cm}X
    S[table-format=2.1]
    S[table-format=1.4]
    S[table-format=1.4]
}
\newcommand{\wtableHeading}{%
  {\bfseries Description} & {\bfseries Acc.} & {\centering\bfseries Attack ARA} & {\centering\bfseries BTR ARA} \\
  \midrule
}

\wtable[inline=, label=tab:results:adv-mods-cifar]
    {Effect of Modifications on CIFAR-10}
    {\wtableColumns}
    {%
\wtableHeading
\tableCifarMods
    }
    {Effects of different modifications from \cref{sec:meth} on the overall classification accuracy and adversarial resistance of an NN classifying CIFAR-10 images.  ``Acc.'' is the accuracy on CIFAR-10's test data after the final epoch.  ``Attack ARA'' and ``BTR ARA'' are both as described in \cref{sec:meth:adv}.  \note{WW}{No weight regularization unless otherwise specified.  Output leveling applied at strength $k_{out} = 0.01$ unless otherwise specified.  Also, check BTR ARA described.}}

\subsubsection{Traditional ResNet-44}
Our baseline ResNet-44 result is denoted in the first row of both \cref{tab:results:adv-mods-cifar,tab:results:adv-adapt-cifar}.  The \textcolor{exp-original}{\tcolordot}\ dot in the row means that it corresponds to the experiment with the same dot in \cref{fig:results:adv-acc,fig:results:adv-compare}.  For this model, none of the adversarial explanations are sensible to a human observer, yet result in a significant change in the network's output (see \cref{app:cifar} for more examples).

\subsubsection{Varied $\psi$ from \cref{eq:meth:lipschitz-loss}}\label{sec:results:cifar:psi}
We sought to verify that increasing the strength of our proposed regularization would lead to an increase in adversarial resistance.  The first section of \cref{tab:results:adv-mods-cifar} demonstrated that this was the case, with the classifier's attack and BTR ARAs increasing monotonically with the strength of the regularization.  Notable also is that, up to a certain level of $\psi = 4.0$, we were able to maintain the classifier's accuracy on clean data while gaining additional adversarial resistance.  After that, clean accuracy decreased as adversarial resistance increased.  Therefore, $\psi$ may be varied in accordance with which is more desirable: accuracy or adversarial resistance.

We also note that the training accuracy never reached \SI{100}{\%} for these experiments, indicating that a ResNet-44 does not have the ability to express a solution to the classification problem which both optimizes accuracy and has derivatives approximately equal to zero; this is explored further in \cref{sec:results:cifar:net-size}.

\subsubsection{Varying $K$ from \cref{eq:meth:lipschitz-loss}}\label{sec:results:cifar:k}
The stochastic formulation of \cref{eq:meth:lipschitz-loss} was expected to yield the same results as a non-stochastic formulation.  To check the validity of this assumption, we tried different values of $K$, from $1$ to $8$.  These experiments demonstrated that varying $K$ had little effect.  As such, all subsequent experiments used $K=1$, which is more efficient to compute than any greater $K$ as it only requires one additional backpropagation per training batch.  

These experiments were virtually identical, with accuracy having a standard deviation of $\pm$\SI{0.2}{\%}, and ARA metrics having a standard deviation of $\pm$\SI{0.0001}{}.  Thus, they demonstrated that training with the proposed regularization produced results with little variance.

\subsubsection{Regularization methods}\label{sec:results:cifar:regularization}
From the prior groups of experiments, it may be concluded that the adversarial resistance loss proposed in \cref{eq:meth:lipschitz-loss} provided a useful form of regularization.  We wanted to test the combination of using \cref{eq:meth:lipschitz-loss} with other regularization techniques.  Due to their promising results on CIFAR-10, we investigated Stochastic Depth \cite{Huang2016stochasticDepth} and ShakeDrop \cite{Yamada2018shakedrop}.  Our experiments with these showed that the proposed regularization performed best on its own, with additional regularizations resulting in lower accuracy on the clean data as well as when dealing with an adversary.  We note that, as our networks' training accuracy on the final epoch were never significantly higher than their validation accuracy, it is likely that a network would need significantly higher capacity before additional regularization would be useful.

\subsubsection{Varied network depth/width}\label{sec:results:cifar:net-size}
As mentioned in \cref{sec:results:cifar:psi,sec:results:cifar:regularization}, the training loss never approached zero when using the proposed regularization.  We assumed this was due to a lack of model capacity.  As such, we tried two different variations of the traditional ResNet, each having roughly $4\times$ as many parameters as the original network.  ResNet-170 is four times as deep, and we also used a ResNet-44 with filters of size $[32, 64, 128]$ for each of the three residual blocks, rather than the traditional $[16, 32, 64]$.  Interestingly, the ResNet-170 did not appear to have an easier time optimizing the training accuracy.  The double width ResNet-44, however, improved slightly in both accuracy and adversarial resistance.  As will be seen in \cref{sec:results:cifar:adapt-net-size}, we found that another trick was required to fully utilize additional network capacity.

At this point, one can begin to see the difference between attack ARA and BTR ARA.  Though the $\psi=12,000$ used for these experiments resulted in a minor increase in attack ARA from \SI{0.0084}{} for $\psi=220$ up to \SI{0.0110}{}, the accuracy fell from \SI{84.5}{\%} down to \SI{56.5}{\%}.  The BTR ARA jumped from \SI{0.0135}{} to \SI{0.0347}{}.  This indicates that while the $\psi=12,000$ classifier was worse at accurately identifying an object in a tiny image, it retained predictive power against adversaries twice as aggressive, and was much better at recognizing features of objects within the dataset.  This manifested as clearer images - many of the adversarial perturbations for $\psi=220$ still looked like randomized noise, whereas the adversarial perturbations for $\psi=12,000$ looked like deliberate changes to the objects in the image.

\subsubsection{``Dead zone'' from \cref{sec:meth:lipschitz}}\label{sec:results:cifar:deadzone}
A true Lipschitz-enforcing loss would not require any penalty on derivatives inside a region $[-\sigma, \sigma]$.  These experiments demonstrated that increasing $\sigma$ results in higher accuracy but less adversarial resistance, particularly in the BTR category.  An additional experiment which used $\sigma=0$ but set $\psi$ such that the final accuracy is about the same demonstrates that there seemed to be no benefit from the addition of this metaparameter, and that leaving it at $\sigma=0$ would seem to be the best choice.  Part of this was due to the difficulty of setting $\sigma$: different values of $\psi$ caused $|\partial y_{n_k} / \partial x_j|$ to be at different scales, and it was difficult to decide on a good value of $\sigma$.  As we will argue in \cref{sec:results:cifar:adapt-ladv}, there is a better way to implement a true Lipschitz constraint in \cref{eq:meth:lipschitz-loss}, should it be beneficial.

\subsubsection{Half-Huber ReLU from \cref{sec:meth:relu2b}}\label{sec:results:cifar:relu2b}
The experiments before this point have used a traditional ReLU; here we used the HHReLU instead, which has a continuous derivative.  At first glance the HHReLU was a modest improvement: improved accuracy and attack ARA, but substantially decreased BTR ARA.  However, by training a ReLU network with an adjusted $\psi$ such that the accuracy was about the same as the HHReLU version of the network, we saw that the HHReLU version of the network was better in both attack and BTR ARAs.  Thus, the HHReLU is an important part of our regularization method, helping networks to learn better structural properties while retaining raw classification accuracy.

We note another potential explanation for a smaller BTR ARA in the first experiment with HHReLU: the HHReLU makes gradients that are easier to follow, and consequently also eases the task of generating successful attacks.  That is, the BTR ARA metric for the ReLU network may be inflated as our adversary was unable to find low-perturbation attacks due to the increased difficulty of following gradients in a ReLU network.

\subsubsection{Varied $\zeta$ from \cref{sec:meth:lipschitz}}\label{sec:results:cifar:zeta}
These results demonstrated that $\zeta > 0$ was capable of inducing better results on the ARA metrics, but at a cost of some accuracy.  We will revisit $\zeta$ in \cref{sec:results:cifar:adapt-zeta}.

\subsubsection{Output zeroing from \cref{sec:meth:outzero}}\label{sec:results:cifar:outzero}
This segment of experiments contained two interesting outcomes.  The first was that biasing the pre-softmax part of the network toward zero via $k_{out} = 0.01$ slightly improved accuracy and slightly worsened adversarial resistance.  The second was that doubling the number of epochs improved accuracy but worsened adversarial resistance.  This was likely due to adversarial overfitting: the $E[|\partial y_{n_k} / \partial x_j|^2]$ of the training data was lower for the double epoch version, but the same quantity for the testing data was higher.  This also made sense with respect to improved accuracy: as the loss from \cref{eq:meth:lipschitz-loss} approached zero, more training bandwidth would be freed up for the classification loss.

Interestingly, when considering an adaptive $\psi$ value, we found that this parameter produced a more pronounced effect: see \cref{sec:results:cifar:adapt-outzero}.

\subsubsection{Weight regularization from \cref{sec:meth:weightreg}}\label{sec:results:cifar:weightreg}
In \cref{sec:meth:weightreg} we proposed that \cref{eq:meth:lipschitz-loss} might be a replacement for the $L_2$-loss traditionally imposed on weights as part of the NN training process.  This experiment demonstrated worse performance without traditional $L_2$ regularization, indicating that architectures deeper than a single layer benefit from both regularization terms.  However, in \cref{sec:results:imagenet:weight}, we elaborate on the need for less $L_2$-regularization when using our technique with large networks.



\wtable[pos=ht!, inline=, label=tab:results:adv-adapt-cifar]
    {Effect of Adaptiveness on CIFAR-10}
    {\wtableColumns}
    {%
\wtableHeading
\tableCifarAdapt
    }
    {Effects of adaptive $\psi$ from \cref{sec:meth:adaptive} on network accuracy and adversarial resistance.  Columns are the same as \cref{tab:results:adv-mods-cifar}.}

\subsubsection{Adaptive $\psi$ from \cref{sec:meth:adaptive}}
All previous experiments were executed with fixed values of $\psi$.  Unfortunately, $\psi$ is a somewhat obtuse parameter, as shown by most of the experiments thus far: it trades between accuracy and ARA in a consistent, but difficult to control manner.  Furthermore, as shown by the double-epochs experiment from \cref{sec:results:cifar:outzero}, fixing its value might be responsible for a type of overfitting.

This group of experiments tested whether targeting a specific training loss -- a known quantity with a more consistent meaning than a specific $\psi$ value -- resulted in any beneficial behavior.  We first took an experiment with known good parameters at a fixed $\psi$, and noted its classification loss on training data for the final epoch: $L = 1.007$.  Setting $L_{target}$ from \cref{eq:meth:adaptive:psi} to this value, the proposed regularization with adaptive $\psi$ resulted in a sizeable accuracy bump, from \SI{78}{\%} to \SI{81}{\%}, while retaining the same adversarial resistance.  Using the same final $\psi$ with a fixed network resulted in similar adversarial resistance, but lower accuracy.

\note{WW}{Clearpage needs to be moved to appropriate point for final draft.}\clearpage
\subsubsection{Varied network depth/width}\label{sec:results:cifar:adapt-net-size}
Revisiting the varied network sizes of \cref{sec:results:cifar:net-size}, but using HHReLU, we compared fixed and adaptive $\psi$ formulations.  Unlike \cref{sec:results:cifar:net-size}, the fixed $\psi$ versions of these larger networks did demonstrate significant improvements to accuracy compared to the baseline ResNet-44, potentially due to HHReLU having increased the quality of the gradients when also contending with the adversarial loss of \cref{eq:meth:lipschitz-loss}.  However, adversarial resistance declined.

For the original ResNet-44 architecture, an adaptive setting of $\psi$ was moderately better for accuracy and worse for ARA metrics.  Recall that this group of experiments used $k_{\psi,0}=0.01, \epsilon_{worse}=1$ from \cref{sec:meth:adaptive}, indicating a very small $\psi$ initially.  We suspected that the change in performance may have been due to the ``shock'' of suddenly adding a new regularization term, and that starting with a larger $k_{\psi,0}$ might alleviate the problem.  To test this, we added an additional experiment with $k_{\psi,0}=220$ and $\epsilon_{worse} = 0.01$ from \cref{sec:meth:adaptive}.  This experiment retained much of the accuracy benefit of using an adaptive $\psi$, while recovering much of the lost adversarial robustness.

For deeper or wider networks, the adaptive $\psi$ versions demonstrated improvements in both accuracy and adversarial ARA over the similar $k_{\psi,0}=0.01$ experiment with a basic ResNet-44.  We theorize that over-penalizing gradients early in network training stymies growth, whereas gradually adding the gradient penalty allows the network to first establish knowledge and subsequently refine it, allowing better usage of additional parameters.  Under this adaptive scheme, the double width network outperformed the quadruple depth network by a narrow margin.

\subsubsection{Different $L_{adv}$ from \cref{sec:meth:lipschitz} with $L_{target} = 1.007$}\label{sec:results:cifar:adapt-ladv}
All previously mentioned experiments were conducted using $L_{adv,z=2}$ from \cref{eq:meth:lipschitz-loss}.  Though the original theory in \cref{sec:meth:lipschitz} indicated $z=1$ would be logical based on the behavior of derivatives in a linear network, actual networks are non-linear.  In prior NN regularization work, the $L_2$ method of weight regularization has consistently been more effective in terms of final accuracy.

This segment of experiments reinforced that $z=1$ performed unambiguously worse than $z=2$.  Interestingly, larger values of $z$ led to increasing amounts of adversarial resistance, at a cost of accuracy.  Note that due to $L_{target}$, all of the different $z$ experiments had similar final training losses, and their testing losses were also all about the same, with a mean and standard deviation of $1.03 \pm 0.02$.  Therefore, the decline in accuracy probably came from increased bias due to the interrelation of the proposed regularization method and the bias/variance trade-off.

We note that large values of $z$ with an adaptive $\psi$ approaches a true Lipschitz constraint at its limit, with a variable constraint on the derivative given by the interaction of $\psi$ and $z$.

We also note that the improved attack ARA performance of $z > 2$ appears to have been unique to CIFAR-10; see \cref{sec:results:imagenet}.  However, BTR ARA increases were consistent on the ILSVRC task as well (\cref{sec:results:imagenet}).

Experiments with $q$ from \cref{sec:meth:lipschitz} were also conducted, and showed similar trade-offs, replacing accuracy with increased ARA.  We leave further exploration of this hyperparameter space to future work.

\subsubsection{$L_{adv,tandem}$ from \cref{sec:meth:lipschitz}}\label{sec:results:cifar:adapt-zeta}
The first two experiments in this section reprised the results from \cref{sec:results:cifar:zeta}, though with an adaptive $\psi$.  The third experiment demonstrated that much of the accuracy loss from $\zeta > 0$ could be recovered by smoothing the difference between the true label and the next-most-confident label (as opposed to smoothing the true label alone).  Furthermore, this technique resulted in higher attack and BTR ARA metrics.

The fourth experiment demonstrates what happened when $L_{adv,tandem}$ was changed to use addition instead of subtraction.  Accuracy improved further, but attack and BTR ARAs dropped significantly.  We have no explanation for that particular phenomenon at this time.

The fifth experiment, with $L_{target}=1.5$, was ran for parity with experiments in \cref{sec:results:cifar:adapt-combined}.  While accuracy slinked down from \SI{80.2}{\%} to \SI{68.7}{\%}, both ARAs increased.  The BTR ARA increased most significantly, from \SI{0.0302}{} to \SI{0.0423}{}.  We note that this is the highest BTR ARA of any of the experiments mentioned thus far, demonstrating that the proposed regularization continues to scale and provide benefits even into substantially decreased levels of accuracy.

\subsubsection{Adversarial / noisy training from \cref{sec:meth:adv-train}}\label{sec:results:cifar:adv-train}
This group of results is divided into four sub-groups: reproducing Madry \etal's results, using Madry \etal's technique with the $L_{2,min}$ adversary, combining our regularization with adversarial training, and combining our regularization with Gaussian noise.

\paragraph{Madry \etal\ method, using $L_2$ adversarial training} 
These experiments used adversarial training as a standalone technique to provide resistance to adversarial examples.  They were modeled off of the prior work of Madry \etal\ \cite{Madry2017towardsResistantAdv} and from personal communication with A. Madry.  The \textcolor{exp-madry}{\tcolordot} dot indicates the N2 experiment labeled in \cref{fig:results:adv-acc,fig:results:adv-compare}, and was a reproduction of the best results from personal communication with A. Madry, albeit with a ResNet-44 instead of a ResNet-50.  This experiment was therefore used for our comparison with the state of the art.

We note that increasing $\epsilon$ from the recommended value of $0.01$ for CIFAR-10 had little beneficial effect: accuracy surprisingly increased, but only from \SI{87.4}{\%} to \SI{88.6}{\%}, and attack ARA decreased from \SI{0.0107}{} to \SI{0.0077}{}.  Interestingly, BTR ARA did increase, from \SI{0.0153}{} to \SI{0.0205}{}.  We hypothesize this was due to BTR ARA measuring the classifier's ability to recognize features of object classes, without penalizing for related classes.  Since the adversarial perturbations were substantially larger, more trucks could be made to look like automobiles, for instance, and the differences between these classes broke down even though the classifier improved at distinguishing them from the other classes such as bird, dog, etc.  This phenomenon was discussed previously in \cref{sec:meth:lipschitz}.

\paragraph{Madry \etal\ method, but with $L_{2,min}$ training}\label{sec:results:cifar:adapt-madry-l2min}
We next tested our proposed $L_{2,min}$ method of adversarial training, described in \cref{sec:meth:adv-train}.  The baseline measurements at $\epsilon=0.01$ were comparable though slightly worse than those for $L_2$ adversarial training.  The measurements at a higher $\epsilon=0.1$ showed drastically decreased accuracy, but significantly higher attack ARA and comparable BTR ARA to the $L_2$ training with the same $\epsilon$.  The decreased accuracy was likely due to $L_{2,min}$ causing a sort of degeneracy, explained in \cref{fig:results:cifar:l2-min-degeneracy}.  Regardless, when combined with Half-Half adversarial training, denoted as ``HHAT'' in the results, $L_{2,min}$ training recovered much of its lost accuracy while retaining the attack ARA and BTR ARA benefits.

\fig[label=fig:results:cifar:l2-min-degeneracy]
    {\graphic{figs/l2-min-degeneracy.svg}}
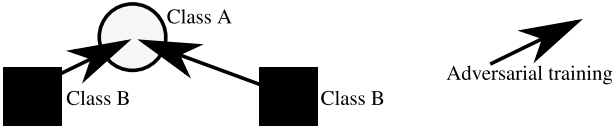
    {The $L_{2,min}$ method of adversarial generation adds stability to adversarial training, but can suffer from degeneracy, where multiple training examples of the same class override a lone neighbor of a different class.  Here, $L_{2,min}$ adversarial training would result in ``Class A'' being overshadowed by the two ``Class B'' instances.  Since $L_{2,min}$ stops at the border of a misclassification, the class with fewer local members only successfully trains a very narrow region.  As per \cref{sec:results:cifar:adapt-madry-l2min}, this may be fixed with HHAT.}

In keeping with the other experiments of \cref{tab:results:adv-adapt-cifar}, many of the adversarial training experiments were conducted with HHReLU rather than ReLU.  This was unlikely to affect the results, as HHReLU is very close to ReLU for approaches that minimize only loss and not its derivatives.  One additional experiment was run to ensure that this did not make a difference.  As predicted, accuracy and ARA statistics are virtually identical for adversarial training with and without HHReLU.

To further test the progression of $L_{2,min}$ into larger values of $\epsilon$, we conducted two further experiments with $\epsilon = 0.25$.  These showed a surprising increase in accuracy for the non-HHAT version, and predictably increased ARA ratings.  The HHAT version followed the expected course of decreased accuracy and increased robustness.  

One aspect we wish to point out is that the best-case attack ARA from these experiments, which used only adversarial training, is on-par or slightly better than the best-case attack ARA using only our proposed regularization, when compared at the same level of accuracy on clean data.  However, the BTR ARA was lower for any of the adversarial training experiments when compared to the BTR ARAs for our regularization.  We therefore posit that adversarial training helps to stabilize the direction of steepest ascent for the loss function, while our proposed regularization stabilizes the entire loss surface.  The definitions of the two techniques provide this distinction, and the empirical evidence appears to support it.

\paragraph{\Cref{eq:meth:lipschitz-loss} with adv. training, using $L_{target}=1.5$ and $L_{adv,tandem}$}\label{sec:results:cifar:adapt-combined}
Combining what was learned from the previous sections, experiments were conducted with a combination of adversarial training and our proposed regularization.  These yielded the best results, improving over the previous bests in both attack ARA and BTR ARA for given levels of accuracy.  The HHAT variety of adversarial training best preserved the benefits to BTR ARA, so that is what we recommend moving forward.  We also note the importance of both $L_{2,min}$ adversarial training and an appropriately high value of $\epsilon$ for attack ARA.  While smaller values of $\epsilon$ still yielded excellent BTR ARA, this was also seen in \cref{sec:results:cifar:adapt-zeta}, and as such likely came almost entirely from our method.

\paragraph{Gaussian noise, using $L_{target} = 1.0$ and $L_{adv,tandem}$} 
The combination of our method with adversarial training was motivated by an attempt to find ``loss cliffs'' (\cref{sec:meth:adv-train}).  To ensure that the computational overhead of adversarial training added value to this cause beyond that of random noise, we also ran several experiments with Gaussian noise added on a per-component basis.  In these experiments, each color value of each pixel received a perturbation independent of all other colors on all other pixels.  Again, the BTR ARA was mostly preserved, but the attack ARA was substantially lower than when combined with adversarial training.

\subsubsection{Combined adversarial training with output zeroing from \cref{sec:meth:outzero}}\label{sec:results:cifar:adapt-outzero}
These two experiments demonstrated that the output zeroing method can have a more significant impact on accuracy without affecting attack ARA when using an adaptive $\psi$, but that the overall benefit was likely not worthwhile, particularly when considering the additional metaparameter.  Nonetheless, most of the experiments in this paper were conducted with output zeroing as in \cref{sec:meth:outzero}.

\subsection{Active Learning from \cref{sec:meth:hitl}}\label{sec:results:cifar:hitl}
Two datasets were investigated for active learning: CIFAR-10 and the JSRT dataset.

In CIFAR-10, two annotators annotated overly-saturated adversarial images generated from a double-width ResNet-44 via the UI in \cref{fig:results:hitl-ui}.  These adversarial images were produced from the CIFAR-10 training data.  One annotator annotated \SI{730}{} and the other annotated \SI{2107}{} images.  Of these annotations, $106$ were of the same images, and $72$ of those were annotated with the same decision (changed, unchanged, or unsure), indicating that annotators agreed on \SI{68}{\%} of the images.  The double-width ResNet-44 architecture was then re-trained from scratch using a dataset consisting of the original CIFAR-10 training data concatenated with the adversarial images annotated ``unchanged.''  The first experiment only used annotations from the first annotator, and had a total of $448$ adversarial images with an ``unchanged'' target class added to the training dataset; the second experiment had a total of \SI{1331}{} images added.  Example adversarial examples annotated as ``changed'' and ``unchanged'' may be found in \cref{app:hitl}.

In the first experiment, adding a small number of annotations made virtually no difference.  Note that CIFAR-10 has \SI{50000}{} training images, so we only increased the dataset's size by \SI{0.9}{\%}.  In the second experiment, which added \SI{2.7}{\%}, we saw an approximately \SI{3}{\%} gain in both attack ARA and BTR ARA, with little change in accuracy.  These gains could potentially be improved by stacking adversarial training with the technique.  A comparison of the adversarial examples from these networks may be found in \cref{app:hitl}.  Given that the examples added to the dataset came from the dataset itself, a linear improvement in attack defense was very promising.  This indicated that a smaller dataset might find more benefit from adding training examples in this manner.

The JSRT dataset consists of only 247 images, of which 199 were used for training.  An annotator annotated \SI{150}{} of the training images using the UI from \cref{fig:results:hitl-ui}, \SI{46}{} of which were marked ``unchanged,'' meaning that the training dataset was increased by \SI{23}{\%}.  A baseline JSRT classification network with no additional training data had an attack ARA of \SI{0.00057}{} on the remaining $48$ testing images, and a modified network trained with the additional annotations had an attack ARA of \SI{0.00095}{}, an increase of $1.7\times$.  To check that this was not a function of a larger training set, we trained an additional network using random annotations, selecting \SI{46}{} of the \SI{150}{} annotations without regard for the annotation.  The network trained with random annotations had an attack ARA of \SI{0.00049}{}, or $0.9\times$ the original network's ARA.  Annotations were therefore very important for improving the network, validating the need for an active learning pipeline.  We note that the $1.7\times$ increase in ARA might be somewhat inflated from calculation noise, as the JSRT ARAs were evaluated on only $48$ testing images.  Nonetheless, contrasting the value of active and random annotations, there was a clear, beneficial effect, which future work might further elucidate.

On datasets with few samples, an active learning pipeline might be a very valuable way to expand training data, assuming the availability of reliable human annotations.

\subsection{ImageNet Experiments}\label{sec:results:imagenet}

Our results on CIFAR-10 were encouraging, but not a guarantee that the technique would extend to larger networks with more complicated tasks.  We trained several networks on ILSVRC 2012, but were somewhat limited in experiments due to each taking roughly a week to train on our hardware without adversarial training, and several weeks with adversarial training.  Nonetheless, we validated our ResNet-18 implementation with a top-1 accuracy of \SI{65.6}{\%}, consistent with literature given we used \SI{65}{} epochs rather than the usual \SI{90}{}.  Results for the following sections are found in \cref{tab:results:adv-mods-imagenet}.

\wtable[label=tab:results:adv-mods-imagenet]
    {Effect of Modifications on ImageNet and COCO}
    {\wtableColumns}
    {%
\wtableHeading
\tableImagenet
    }
    {Effects of different modifications from \cref{sec:meth} on the overall classification accuracy and adversarial resistance of an NN classifying ImageNet and COCO images.
    \\ $\dagger$ This experiment was aborted after 18 epochs as it went unstable; the experiment below it had an accuracy of \SI{43.0}{\%} at the same number of epochs.}

\subsubsection{Weight regularization from \cref{sec:meth:weightreg}}\label{sec:results:imagenet:weight}
While using the proposed regularization to replace $L_2$ weight regularization as per \cref{sec:meth:weightreg} did not pan out for CIFAR-10, in ImageNet we found that it was vital to reduce the amount of $L_2$ weight regularization from $1e-4$ to $1e-6$ for the network to converge with our regularization.  This was not required for the standard ResNet-18.  Relative to the task, ResNet-18 is likely underparameterized, and the classification loss, $L_2$ loss, and proposed $L_{adv}$ loss were likely too at odds to find a good solution.  Reducing the amount of $L_2$ loss made the problem tractable again.

\subsubsection{Automatic $\psi$ from \cref{sec:meth:adaptive}}
Given its efficacy on CIFAR-10, and that it was a better parameterization of the problem, we conducted all but one of the ILSVRC 2012 experiments with an adaptive $\psi$.  All of these were conducted with HHReLU, an $L_2$ weight regularization of 1e-6, no dead zone, $\sigma = 0$, $k_{out} = 5e-5$, and $L_{adv,z=2}$ unless otherwise specified.  The first two experiments in this group showed that, again, an adaptive $\psi$ outperformed a fixed value of $\psi$.  The subsequent experiments demonstrated the existence of the accuracy/attack RMSE trade-off, just like with CIFAR-10.  However, with \SI{1000}{} classes, ImageNet's attack ARA did not scale well as accuracy fell.  The BTR ARA scaled well.  See \cref{app:imagenet} for examples of adversarial examples generated on these networks.

The proposed regularization method worked well on ILSVRC 2012, and was capable of generating convincing adversarial examples for many of the target classes.  Other target classes, such as ``n01484850 great white shark,'' were clearly underspecified in the dataset, probably due to a lack of other classes with similar features.  Many shark images are predominantly water, a property shared by few other ILSVRC 2012 classes.  Similarly, the adversarial explanations resulted in the addition of water to the input more than any other feature.

When comparing this ARA with that of Cohen \etal's \cite{Cohen2019certified}, we point out that our computational resources only allowed for a ResNet-18 on the ImageNet challenge versus their ResNet-50.  The literature around this topic, and our own work when dealing with an adaptive $\psi$, supports that larger networks tend to demonstrate greater adversarial resistance \cite{Cohen2019certified,Tsipras2018robustness}.

%
%
%

\subsection{COCO Experiments}\label{sec:results:coco}

Experiments were conducted on COCO to determine the efficacy of our regularization.  As described in \cref{sec:meth:arch:coco}, the COCO dataset was a somewhat unique experiment as many of the images overlapped with other classes and the ``person'' class was over-represented as \SI{31.4}{\%} of the total dataset.

Without any methods providing adversarial resistance, our COCO network scored \SI{77.3}{\%} accuracy with an attack ARA rating of \SI{0.0003}{}.  Note that the high accuracy of a naive classifier -- \SI{31.4}{\%} -- swallows up much of the area that would otherwise increase the ARA ratings on this problem.  Adversarial training added a good amount of attack ARA but only a little BTR ARA, consistent with previous experiments from \cref{sec:results:cifar:adapt-madry-l2min}.  However, using only our technique without any adversarial training resulted in the best overall statistics.  We initially supposed this was due to the class imbalance, as $L_{2,min}$ adversarial training can suppress the correct label (\cref{fig:results:cifar:l2-min-degeneracy}).  Unfortunately, experiments with balanced training on the COCO dataset, such that the classification loss for each label was divided by the percentage of that label, still resulted in little benefit from adversarial training.  It is thus very possible that adversarial training did not make the COCO networks more robust as a consequence of the high number of overlapping objects in different frames --- the actual distances between classes in the base problem were sufficiently small that adversarial training offered little benefit.

\section{Conclusion}\label{sec:conc}
We demonstrated a regularization technique based on the Lipschitz constraint, which significantly enhanced the ability of networks to resist adversarial examples.  This was paired with other innovations, including a Half-Huber Rectified Linear Unit and improved adversarial training methodology.  On ILSVRC 2012, the methods in this work increased the ARA by \valImagenetOurBenefitAuc\ over the previous state of the art, while retaining the same level of accuracy on clean data and using a network one-third of the size of the previous state of the art.  More central to the tenets of this work, we demonstrated that the stability added by these techniques allows for adversarial examples to be generated with very discernible features.  These adversarial examples could then be used as non-linear explanation mechanisms, termed adversarial explanations, working with the network and its non-linearities to produce more reliable explanations than prior work.  Furthermore, we demonstrated that AEs might be annotated and fed back into the training process as part of an active learning pipeline to yield improved adversarial resistance.  We hope that this work provides a basis for future work in the realms of both adversarial resistance and explainable machine learning, making algorithms more reliable for industry fields where accountability matters, such as biomedical or autonomous vehicles.



\section*{Acknowledgements}
We would like to thank A. Madry (from \cite{Madry2017towardsResistantAdv,Tsipras2018robustness}) and J. Cohen (from \cite{Cohen2019certified}) for helpful discussions and clarifications about their work.  We thank FuR and Alex Parise for assisting with the collection of photos for the examples throughout this work.

\section*{Author Contributions}
W.W. contributed the original idea, algorithms, experiment design, ablation studies, some active learning annotations, and wrote the majority of the paper.  J.C. contributed LIME and Grad-CAM integrations, annotated the majority of the active learning annotations, provided text for the active learning sections of the paper, and contributed editing support.  C.T. contributed scope advisement, editing support, and funding for the work.

\bibliographystyle{./sty/ieee/IEEEtran-nomonth}
\bibliography{./sty/ieee/IEEEabrv,biblio}

\newif\ifsupplementaltoo
\supplementaltootrue

\ifsupplementaltoo
\clearpage
\onecolumn
\appendix

\setcounter{figure}{0}
\makeatletter
\renewcommand{\thefigure}{S\@arabic\c@figure}
\makeatother

\newcommand{\rulesep}{\unskip\ \vrule\ }



\begin{multicols}{2}
  \multicolslines
\subsection{CIFAR-10 Attack ARA Comparison}\label{app:attack-ara}

This appendix contains adversarial examples against CIFAR-10 networks with differing attack ARAs.  The networks used were the same ones annotated in \cref{fig:results:adv-acc}.  N1 denotes a traditional NN, N2 denotes an NN with only adversarial training, N3 denotes an NN with both our proposed regularization and adversarial training, and N4 denotes an NN trained only with our proposed regularization.  The attack ARAs of N1 through N4 were $0.0013$, $0.0107$, $0.0197$, and $0.0151$, respectively.

The adversarial examples shown in \cref{app:cifar-ara-1,app:cifar-ara-2} were the closest examples to the original images which resulted in a misclassification, found across 450 steps of optimization, as per \cref{sec:meth:adv:accuracy}.  The ``Attacked'' column indicates a misclassified image, the ``Input'' column indicates the original input, and the ``Noise'' column indicates the difference between the two.  If the ``Noise'' column is entirely black, the RMSE will be $0$, indicating that the network was incorrect without any adversarial perturbation.

The most extreme perturbations are found in \cref{app:cifar-ara-1:a}.  In the examples for N2, the adversarially trained network, the changes are barely perceptible, while the N3 network required significant changes for misclassifications to occur.  However, the adversarial examples which were not among the largest perturbations, found in \cref{app:cifar-ara-2}, are largely imperceptible for all networks.  We point this out to motivate future work in increasing these margins, and to point out that while the \SI{84}{\%} increase of attack ARA does set a new state of the art, the most significant gains from our techniques came from illuminating salient features through AEs.
\end{multicols}

\fig[pos=h!, label=app:cifar-ara-1]
  {%
    \subfig[label=app:cifar-ara-1:a, width=0.8\linewidth]{\graphic{plots/app-ara-compare-0.pdf}}
    \subfig[label=app:cifar-ara-1:b, width=0.8\linewidth]{\graphic{plots/app-ara-compare-1.pdf}}
  }
  {See \cref{app:attack-ara} for details.}

\fig[pos=h!, label=app:cifar-ara-2]
  {%
    \subfig[label=app:cifar-ara-2:a, width=0.9\linewidth]{\graphic{plots/app-ara-compare-2.pdf}}
    \subfig[label=app:cifar-ara-2:b, width=0.9\linewidth]{\graphic{plots/app-ara-compare-3.pdf}}
    \subfig[label=app:cifar-ara-2:c, width=0.9\linewidth]{\graphic{plots/app-ara-compare-4.pdf}}
  }
  {See \cref{app:attack-ara} for details.}

\clearpage
\begin{multicols}{2}
\multicolslines
\subsection{CIFAR-10}\label{app:cifar}

This appendix contains example adversarial images from the CIFAR-10 dataset, using the same networks annotated in \cref{fig:results:adv-acc}.  N1 denotes a traditional NN, N2 denotes an NN with only adversarial training, N3 denotes an NN with both our proposed regularization and adversarial training, and N4 denotes an NN trained only with our proposed regularization.  At the top of each column is a label for the target of $g_{explain+}$, applied with $\rho=0.1$.

As per \cref{tab:results:adv-adapt-cifar}, the BTR ARAs of N1 through N4 were $0.0014$, $0.0153$, $0.0450$, and $0.0423$, respectively.  N3 was the best performer in the BTR ARA category.  The below \cref{app:cifar-1,app:cifar-2,app:cifar-3} support the ranking given by the BTR ARA.  For example, in \cref{app:cifar-1:a}, consider which network gave the most compelling explanation for each of the ten categories; we propose N3, N3, N4, N2, N3, N3, N3, N3, N3, and N4, respectively for each target column.  By that count, N3 produced that most compelling explanation \SI{70}{\%} of the time.  Further study of the relative explanatory benefit of these techniques from the subjective view of human operators is merited, but these results indicate that the BTR ARA is a strong measure of explanation quality.

\Cref{app:cifar-4} demonstrates results of applying $g_{explain+}$ with varying levels of $\rho$; see the figure caption for more details.
\end{multicols}

\fig[pos=h!, label=app:cifar-1]
  {%
    \subfig[label=app:cifar-1:a]{\graphic{plots/app_cifar_0.pdf}}
    \subfig[label=app:cifar-1:b]{\graphic{plots/app_cifar_8.pdf}}
    \vspace*{-1em}
  }
  {See \cref{app:cifar} for details.}

\fig[pos=h!, label=app:cifar-4]
  {%
    \subfig[label=app:cifar-4:a]{\graphic{plots/app_cifar_morph_0.pdf}}
    \subfig[label=app:cifar-4:b]{\graphic{plots/app_cifar_morph_1.pdf}}
  }
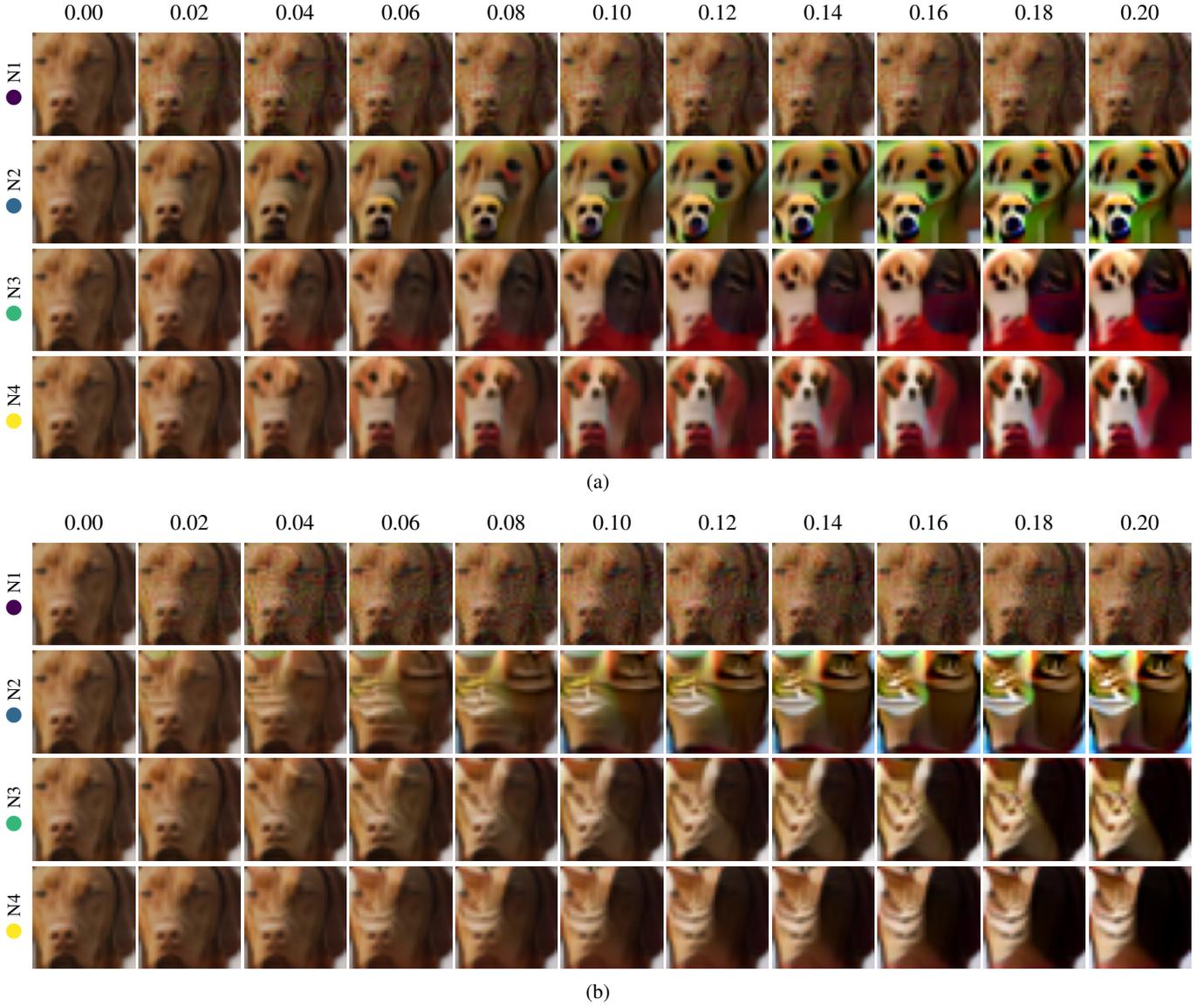
  {While \cref{app:cifar-1} demonstrates the relative performance of different visual explanations for various networks at a fixed noise magnitude $\rho=0.1$, this figure focuses on demonstrating the evolution of those explanations as $\rho$ is varied between $0$ and $0.2$ (the numbers at the top of each column).  The input figure used was the same as \cref{app:cifar-1:b}.  For \cref{app:cifar-4:a}, the target of $g_{explain+}$ is the ``dog'' class, and for \cref{app:cifar-4:b}, the target is the ``cat'' class.  The ``cat'' class was chosen because it was one of the wrong predictions with a higher level of confidence, particularly for the N2 network.  Note that the progression is non-linear, with new features appearing at different levels of $\rho$. \\\indent
    Investigating \cref{app:cifar-4:a}, one can see that the adversarially-trained N2 network relies on a small, dog-like feature around the nostrils of the original input image.  N3 and N4, the networks which additionally have the regularization of \cref{eq:meth:lipschitz-loss} (and the other tricks from \cref{sec:meth}), rely more on the overall shading of the face, and a larger dog-like feature which emerges by reshaping the left side of the input.  From this, one might infer that the majority of CIFAR-10 training images are of full dogs, and the network has adopted this bias. \\\indent
  In \cref{app:cifar-4:b}, the progressions for the N3 and N4 networks make it clear that the shading on the left side of the face may be adapted into the form of a cat, a likely reason for the misclassification.  The N2 network exploits less of the source image to make this happen, but appears to suffer from a similar misconception.}

\fig[pos=h!, label=app:cifar-2]
  {%
    \subfig{\graphic{plots/app_cifar_3.pdf}}
    \subfig{\graphic{plots/app_cifar_4.pdf}}
    \subfig{\graphic{plots/app_cifar_5.pdf}}
  }
  {See \cref{app:cifar} for details.}

\fig[pos=h!, label=app:cifar-3]
  {%
    \subfig{\graphic{plots/app_cifar_6.pdf}}
    \subfig{\graphic{plots/app_cifar_7.pdf}}
    \subfig{\graphic{plots/app_cifar_2.pdf}}
  }
  {See \cref{app:cifar} for details.}

\clearpage
\begin{multicols}{2}
\multicolslines
\subsection{CIFAR-10 Active Learning}\label{app:hitl}

This appendix contains example adversarial images from the CIFAR-10 dataset that were part of the active learning experiments from \cref{sec:results:cifar:hitl}.  \Cref{app:hitl-annot} contains examples from networks trained on the original CIFAR-10 data; \cref{app:hitl-train} contains comparison images between adversarial examples from the original network and versions trained with annotated data.

\subsubsection{Sample Annotations}\label{app:hitl-annot}
\Cref{app:hitl-annot-both1} demonstrates annotations from an annotator.  Images are arranged in pairs; on the left is the original image, and on the right is an adversarial example constructed such that the difference between the adversarial class and the true class is $s_{adv} - s_t=0.5$.  The original class is written below each image.  Annotators were asked whether or not the adversarial image still belonged to the original class; see \cref{fig:results:hitl-ui}.  The values from each annotator are shown following the original label (as ``yes'' or ``no'' to being the same class).
\end{multicols}


\fig[pos=h!, label=app:hitl-annot-both1]
  {\graphic{plots/hitl-annot-a.pdf}}
  {See \cref{app:hitl-annot} for details.}

\clearpage
\begin{multicols}{2}
\multicolslines
\subsubsection{Networks Trained with Active Learning Feedback}\label{app:hitl-train}
\Cref{app:hitl-trained} contains groups of three rows: the version of the network without active learning, from which annotations were made, an active learning network trained with \SI{448}{} additional annotations, and an active learning network trained with \SI{1331}{} additional annotations.  Just as the network with the most annotations demonstrated a modest quantitative improvement (\cref{tab:results:adv-adapt-cifar}), inspection of these images demonstrates a more fully formed idea of each class.  The left-most image of each group is the input, followed by adversarial examples formed from $g_{explain+}$ targeting the classes of airplane, automobile, bird, and cat.

\end{multicols}
\newcommand{\figChopLabels}[1]{}
\newlength{\fw}
\setlength{\fw}{0.49\linewidth}
\fig[pos=h!, label=app:hitl-trained]
  {%
    \subfig[width=\fw]{\graphic{plots/ex_hitl_0.pdf}}
    \subfig[width=\fw]{\graphic{plots/ex_hitl_1.pdf}}
    \subfig[width=\fw]{\graphic{plots/ex_hitl_2.pdf}}
    \subfig[width=\fw]{\graphic{plots/ex_hitl_3.pdf}}
    \subfig[width=\fw]{\graphic{plots/ex_hitl_4.pdf}}
    \subfig[width=\fw]{\graphic{plots/ex_hitl_5.pdf}}
  }
  {See \cref{app:hitl-train} for details.}

\clearpage
\begin{multicols}{2}
\multicolslines
\subsection{COCO}\label{app:coco}
This appendix contains example adversarial images from the COCO dataset.  The networks used in these figures correspond to networks from \cref{tab:results:adv-mods-imagenet}: C1 is the row labeled ``With \cref{eq:meth:lipschitz-loss},'' C2 is the row labeled ``Combined \cref{eq:meth:lipschitz-loss} + HHAT,'' C3 is the row labeled ``Balanced classes, \cref{eq:meth:lipschitz-loss} only,'' and C4 is the row labeled ``Balanced classes, \cref{eq:meth:lipschitz-loss} + HHAT.''  The BTR ARAs for C1 through C4 were $0.0278$, $0.0250$, $0.0335$, and $0.0325$; note that C3 and C4's BTR ARAs used a lower naive baseline for the ARA calculation, as these datasets were balanced. ``<guess>'' is used to indicate an explanation for the highest-confidence prediction which was of the incorrect class, and ``<real>'' is an explanation of the correct class.  The remaining columns show each network's interpretation of the image as the label at the top of each column (categories were chosen for an even distribution over object type).  As described in \cref{sec:results:coco}, these images corroborate that adversarial training had little benefit for the COCO problem, potentially due to the many overlapping objects in the training data.  As might be expected, the networks that were ``balanced'' resulted in less coherent explanations of the ``person'' class but better explanations for the other classes.
\end{multicols}

\fig[pos=h!]
  {%
    \subfig{\graphic{plots/app_coco_0.pdf}}
    \subfig{\graphic{plots/app_coco_1.pdf}}
  }
  {See \cref{app:coco} for details.}
\fig[pos=h!]
  {%
    \subfig{\graphic{plots/app_coco_2.pdf}}
    \subfig{\graphic{plots/app_coco_3.pdf}}
    \subfig{\graphic{plots/app_coco_7.pdf}}
  }
  {See \cref{app:coco} for details.}

\clearpage
\begin{multicols}{2}
\multicolslines
\subsection{ILSVRC 2012}\label{app:imagenet}
These images contain adversarial explanations for networks from \cref{tab:results:adv-mods-imagenet}: I1 is the network labeled ``$L_{target}=3.1$,'' which had no adversarial training, I2 is the network labeled ``$L_{target}=3.1$ with HHAT, $\epsilon=0.1$,'' I3 is the network labeled ``$L_{target}=5.0$,'' and I4 is the network labeled ``$L_{target}=5.0$, $L_{adv,z=2,q=1}$.''  The BTR ARAs for I1 through I4 were $0.0182$, $0.0282$, $0.0388$, and $0.0538$.

Like \cref{app:cifar-4}, the \cref{app:imagenet-1} demonstrates how explanations progress as $\rho$ is varied.  Between I1 and I2, adversarial training was clearly beneficial on ILSVRC, consistent with the BTR ARA results in \cref{tab:results:adv-mods-imagenet}.  We draw particular attention to the diminished effect on the background of the image in \cref{app:imagenet-1:a}, which is lettering on a sign.  While I1, I3, and I4 all add a green hue to the background, the only network trained with both \cref{eq:meth:lipschitz-loss} and adversarial training did not exhibit this effect.  For both \cref{app:imagenet-1:a,app:imagenet-1:b}, the higher the value of $L_{target}$, the simpler and more coherent the explanation.  While I4 exhibited the highest BTR ARA of \SI{0.0538}{} compared to \SI{0.0282}{} for I2, the best attack ARA was from I2, at a value of \SI{0.0053}{} versus \SI{0.0037}{} for I4.

\Cref{app:imagenet-2,app:imagenet-3,app:imagenet-4} follow a similar format to those in \cref{app:coco}, with ``<guess>'' and ``<real>'' having the same meaning, and the remaining columns being various targets for $g_{explain+}$.
\end{multicols}

\fig[pos=h!,label=app:imagenet-1]
  {%
    \subfig[label=app:imagenet-1:a]{\graphic{plots/app_imagenet_morph_0.pdf}}
    \subfig[label=app:imagenet-1:b]{\graphic{plots/app_imagenet_morph_1.pdf}}
  }
  {See \cref{app:imagenet} for details.}
\fig[pos=h!, label=app:imagenet-2]
  {%
    \subfig{\graphic{plots/app_imagenet_0.pdf}}
    \subfig{\graphic{plots/app_imagenet_1.pdf}}
    \subfig{\graphic{plots/app_imagenet_2.pdf}}
  }
  {See \cref{app:imagenet} for details.}
\fig[pos=h!, label=app:imagenet-3]
  {%
    \subfig{\graphic{plots/app_imagenet_3.pdf}}
    \subfig{\graphic{plots/app_imagenet_4.pdf}}
    \subfig{\graphic{plots/app_imagenet_5.pdf}}
  }
  {See \cref{app:imagenet} for details.}
\fig[pos=h!, label=app:imagenet-4]
  {%
    \subfig{\graphic{plots/app_imagenet_9.pdf}}
    \subfig{\graphic{plots/app_imagenet_10.pdf}}
    \subfig{\graphic{plots/app_imagenet_11.pdf}}
  }
  {See \cref{app:imagenet} for details.}

\fi

\end{document}